%% file: neurips_2024.tex
\documentclass{article}

\PassOptionsToPackage{numbers, compress}{natbib}

\usepackage[final]{neurips_2024}

\usepackage[utf8]{inputenc} 
\usepackage[T1]{fontenc}    
\usepackage{url}            
\usepackage{booktabs}       
\usepackage{amsfonts}       
\usepackage{nicefrac}       
\usepackage{microtype}      
\usepackage{xcolor}         
\usepackage{graphicx}
\usepackage{booktabs}
\usepackage{algorithm}
\usepackage{algpseudocode}
\usepackage{courier}
\usepackage{listings}
\usepackage{multirow}
\usepackage{amsmath}
\input{sec/figures}
\definecolor{citeblue}{RGB}{48,111,186}
\usepackage[pagebackref=false,breaklinks=true,colorlinks=true,citecolor=citeblue,bookmarks=false]{hyperref}
\hypersetup{
    colorlinks=true,
    urlcolor=magenta,
}
\title{InterControl: Zero-shot Human Interaction Generation by Controlling Every Joint}

\author{%
Zhenzhi Wang$^1$, 
Jingbo Wang$^2$, 
Yixuan Li$^1$, 
Dahua Lin$^{1,2}$, 
Bo Dai$^{3,2}$ \\
$^1$The Chinese University of Hong Kong,
$^2$Shanghai Artificial Intelligence Laboratory, \\
$^3$The University of Hong Kong \\
  \texttt{\{wz122,ly122,dhlin\}@ie.cuhk.edu.hk, wangjingbo@pjlab.org.cn} \\ 
  \texttt{bdai@hku.hk} \\
}
\begin{document}

\maketitle

\input{sec/0_abstract}

\figTEASER
\input{sec/1_intro}
\input{sec/2_related}
\input{sec/3_method}

\input{sec/4_experiment}
\input{sec/5_conclusion}

\noindent{\bf Acknowledgment.} This project is funded in part by Shanghai Artificial Intelligence Laboratory, CUHK Interdisciplinary AI Research Institute, and the Centre for Perceptual and Interactive Intelligence (CPII) Ltd under the Innovation and Technology Commission (ITC)'s InnoHK. We would like to thank Tianfan Xue for his insightful discussion.
{
\small
\bibliographystyle{ieeenat_fullname}
\bibliography{main}
}
\clearpage
\appendix
\section*{Appendix}
\input{sec/supp}

\clearpage
\section*{NeurIPS Paper Checklist}

\begin{enumerate}

\item {\bf Claims}
    \item[] Question: Do the main claims made in the abstract and introduction accurately reflect the paper's contributions and scope?
    \item[] Answer:  \answerYes{} 
    \item[] Justification: We are the first method to perform zero-shot human interaction generation by leveraging only single-person motion generation model, which could be supported by abstract, introduction and method.

\item {\bf Limitations}
    \item[] Question: Does the paper discuss the limitations of the work performed by the authors?
    \item[] Answer:  \answerYes{}  
    \item[] Justification: We have discussed our limitation in Sec. Conclusion and Limitations. The main limitation is that we only investigated a certain form of interactions which could be quantitatively described by spatial relations.

\item {\bf Theory Assumptions and Proofs}
    \item[] Question: For each theoretical result, does the paper provide the full set of assumptions and a complete (and correct) proof?
    \item[] Answer: \answerNA{} 
    \item[] Justification: We do not have theoretical result.

    \item {\bf Experimental Result Reproducibility}
    \item[] Question: Does the paper fully disclose all the information needed to reproduce the main experimental results of the paper to the extent that it affects the main claims and/or conclusions of the paper (regardless of whether the code and data are provided or not)?
    \item[] Answer: \answerYes{} 
    \item[] Justification: We have included all the information to reproduce the main experimental results, and we also provide the code.

\item {\bf Open access to data and code}
    \item[] Question: Does the paper provide open access to the data and code, with sufficient instructions to faithfully reproduce the main experimental results, as described in supplemental material?
    \item[] Answer: \answerYes{} 
    \item[] Justification: We have provided the code in the supplemental material and the website.

\item {\bf Experimental Setting/Details}
    \item[] Question: Does the paper specify all the training and test details (e.g., data splits, hyperparameters, how they were chosen, type of optimizer, etc.) necessary to understand the results?
    \item[] Answer: \answerYes{} 
    \item[] Justification: We include all training and test details in the appendix.

\item {\bf Experiment Statistical Significance}
    \item[] Question: Does the paper report error bars suitably and correctly defined or other appropriate information about the statistical significance of the experiments?
    \item[] Answer: \answerNo{} 
    \item[] Justification: It will be too computationally expensive.

\item {\bf Experiments Compute Resources}
    \item[] Question: For each experiment, does the paper provide sufficient information on the computer resources (type of compute workers, memory, time of execution) needed to reproduce the experiments?
    \item[] Answer: \answerYes{} 
    \item[] Justification: We include the information of computer resources in appendix.
    
\item {\bf Code Of Ethics}
    \item[] Question: Does the research conducted in the paper conform, in every respect, with the NeurIPS Code of Ethics \url{https://neurips.cc/public/EthicsGuidelines}?
    \item[] Answer: \answerYes{} 
    \item[] Justification: Our paper follows the NeurIPS Code of Ethics.

\item {\bf Broader Impacts}
    \item[] Question: Does the paper discuss both potential positive societal impacts and negative societal impacts of the work performed?
    \item[] Answer: \answerNA{} 
    \item[] Justification: We have carefully considered potential societal impacts and determined that our technical contribution of generating 3D skeletal animations poses minimal risks. Our method is designed for generating multi-people 3D skeletons, and these skeletal representations do not pose negative societal impacts.
    
\item {\bf Safeguards}
    \item[] Question: Does the paper describe safeguards that have been put in place for responsible release of data or models that have a high risk for misuse (e.g., pretrained language models, image generators, or scraped datasets)?
    \item[] Answer: \answerNA{} 
    \item[] Justification: Our work does not pose risks requiring such safeguards.

\item {\bf Licenses for existing assets}
    \item[] Question: Are the creators or original owners of assets (e.g., code, data, models), used in the paper, properly credited and are the license and terms of use explicitly mentioned and properly respected?
    \item[] Answer: \answerYes{} 
    \item[] Justification: We have cited the used code, data and models.

\item {\bf New Assets}
    \item[] Question: Are new assets introduced in the paper well documented and is the documentation provided alongside the assets?
    \item[] Answer: \answerYes{} 
    \item[] Justification: We have included the documentation of our code in the supplementary materials.

\item {\bf Crowdsourcing and Research with Human Subjects}
    \item[] Question: For crowdsourcing experiments and research with human subjects, does the paper include the full text of instructions given to participants and screenshots, if applicable, as well as details about compensation (if any)? 
    \item[] Answer: \answerNA{} 
    \item[] Justification: This paper does not involve crowdsourcing nor research with human subjects.

\item {\bf Institutional Review Board (IRB) Approvals or Equivalent for Research with Human Subjects}
    \item[] Question: Does the paper describe potential risks incurred by study participants, whether such risks were disclosed to the subjects, and whether Institutional Review Board (IRB) approvals (or an equivalent approval/review based on the requirements of your country or institution) were obtained?
    \item[] Answer:  \answerNA{} 
    \item[] Justification: This paper does not involve crowdsourcing nor research with human subjects.

\end{enumerate}
\end{document}

%% file: sec/figures.tex
\newcommand{\figTEASER}{
    \begin{figure}[t]
        \begin{center}
        \includegraphics[width=0.9\linewidth]{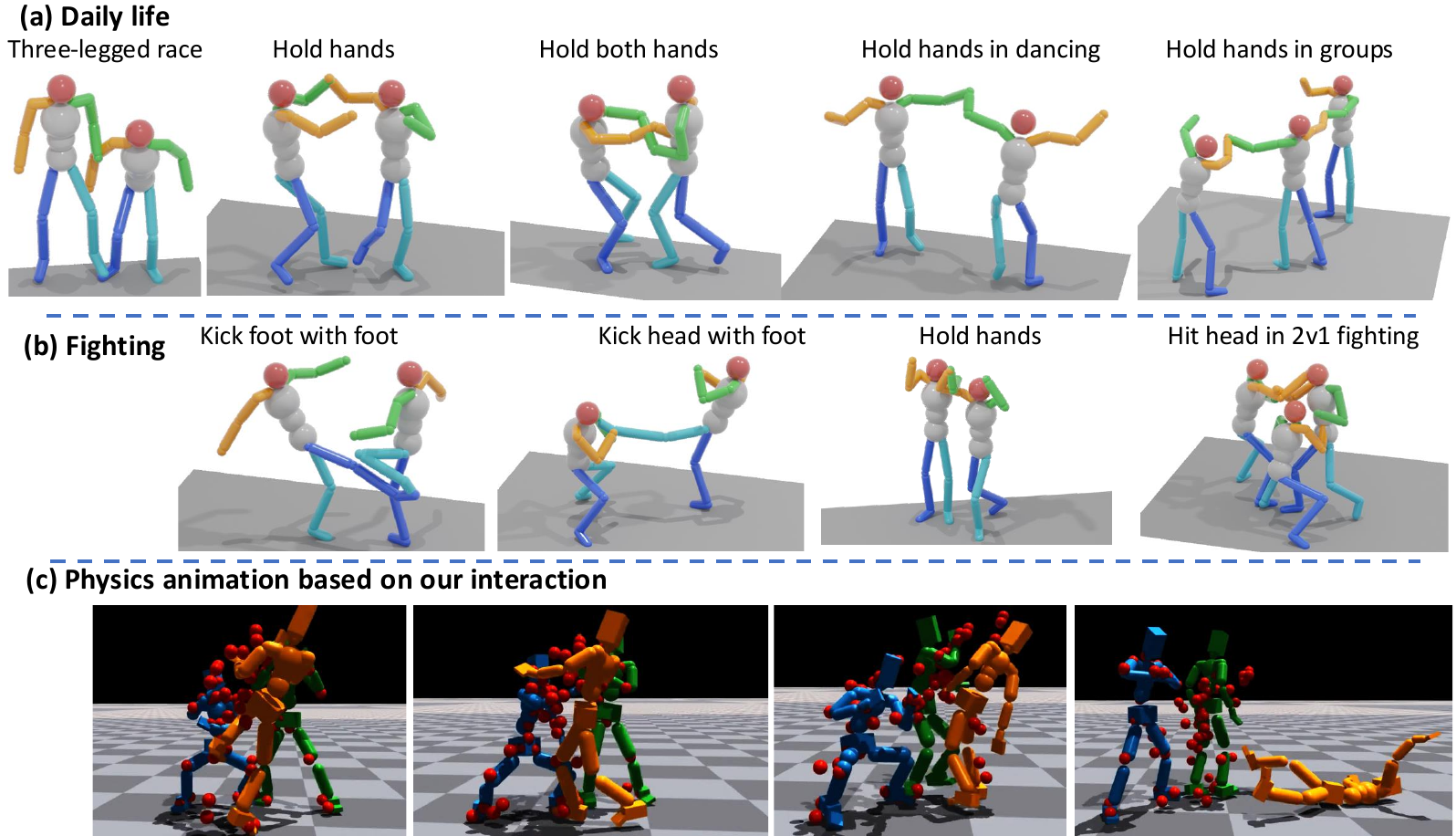}
         \caption{InterControl is able to generate {\bf interactions of a group of people} given joint-joint {\bf contact} or {\bf separation} pairs as spatial condition, and it is only trained on {\bf single-person data}. Our generated interactions are realistic and similar to real interactions in internet images in (a) daily life and (b) fighting. (c) shows our generated group motions (red dots) could serve as reference motions for physics animation.} 
        \label{fig:teaser}
        \end{center}
        \vspace{-1em}
    \end{figure}
}

\newcommand{\figOVERVIEW}{
    \begin{figure*}[t]
        \begin{center}
        \includegraphics[width=1.0\linewidth]{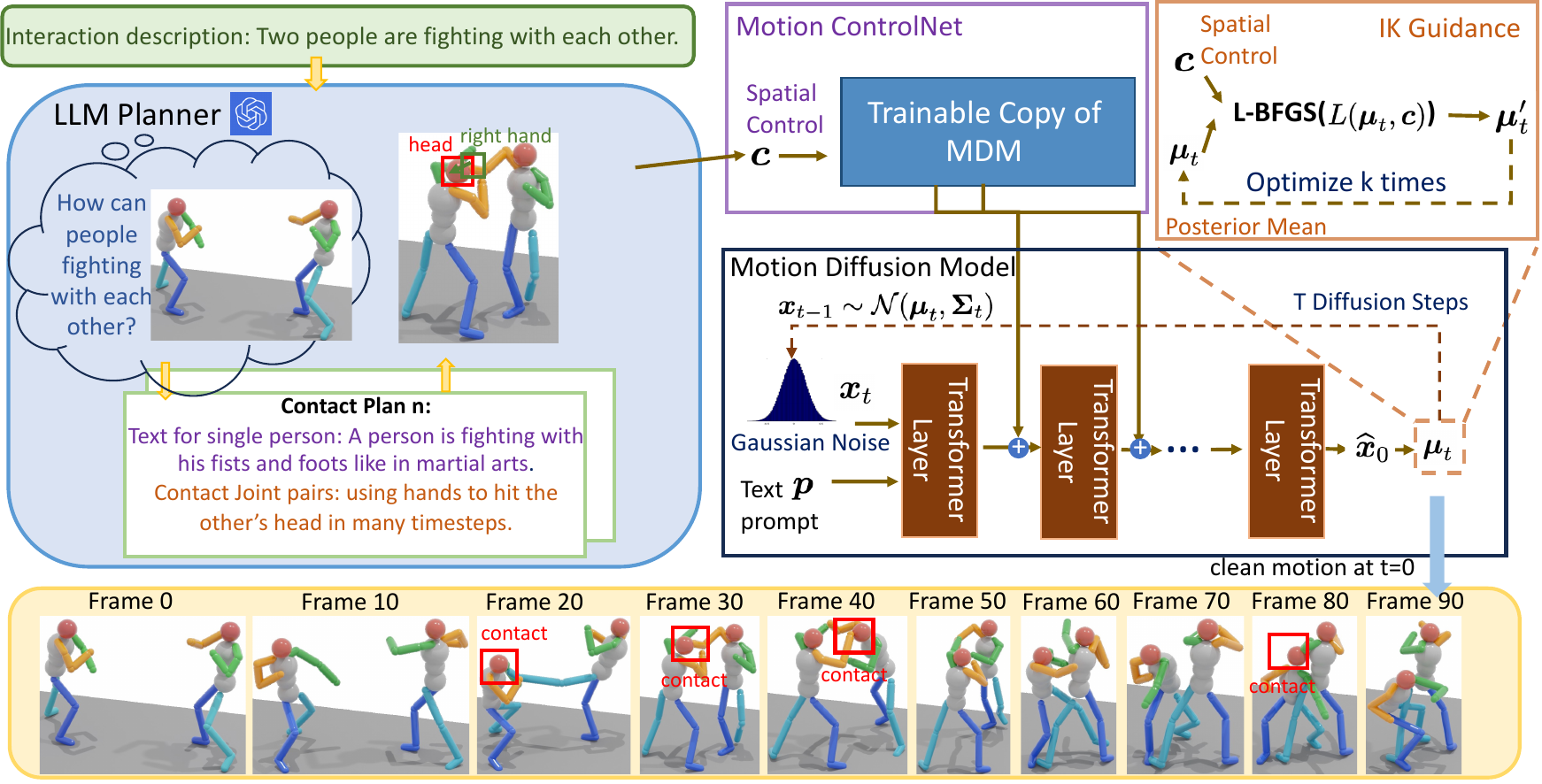}
        \end{center}
        \vspace{-1em}
           \caption{{\bf Overview.} Our model could precisely control human joints in the global space via the Motion ControlNet and IK guidance module. By leveraging LLM to adapt interaction descriptions to joint contact pairs, it could generate multi-person interactions via a single-person motion generation model in a zero-shot manner.}
        \label{fig:framework}
        \vspace{-1em}
    \end{figure*}
}

\newcommand{\figVisual}{
    \begin{figure*}[t]
        \begin{center}
        \includegraphics[width=1.0\linewidth]{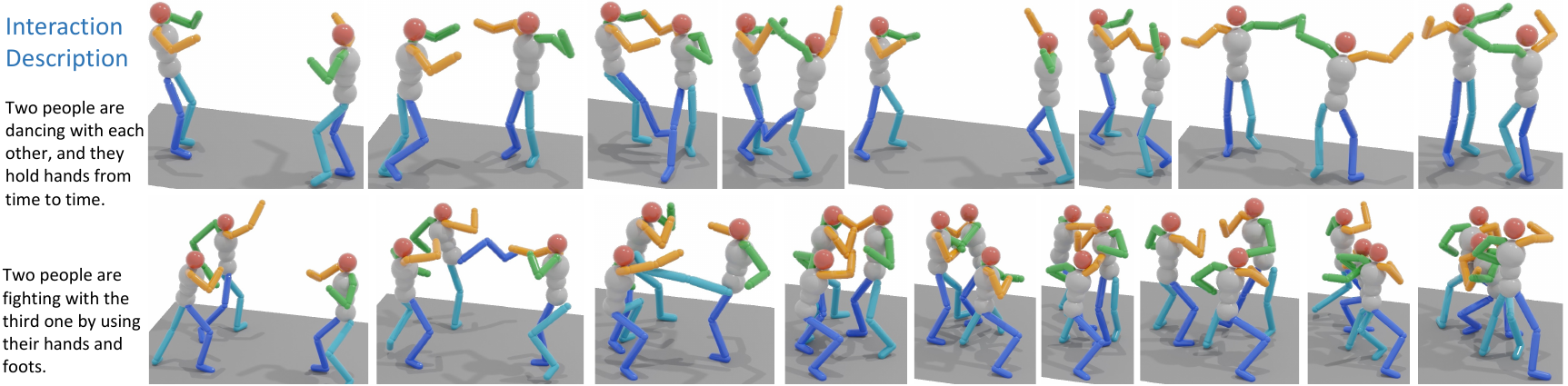}
        \end{center}
        \vspace{-1em}
           \caption{{\bf Qualitative results} of zero-shot human interaction generation. }
        \label{fig:visual-skeleton}
        \vspace{-1em}
    \end{figure*}
}

\newcommand{\figUserStudy}{
    \begin{figure*}[t]
        \begin{center}
        \includegraphics[width=1.0\linewidth]{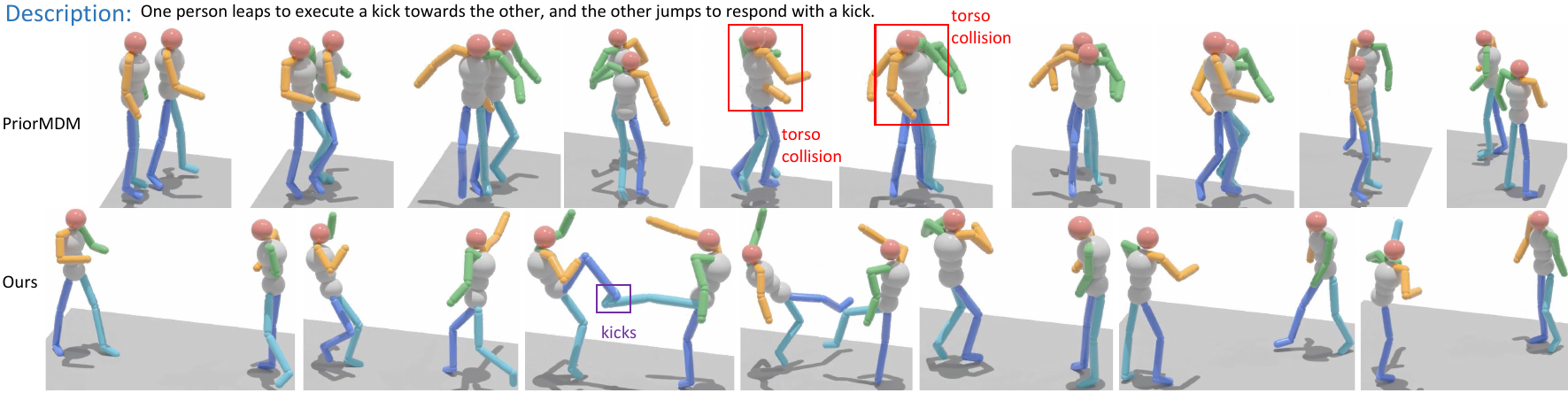}
        \end{center}
        \vspace{-1em}
           \caption{Comparison with PriorMDM~\cite{shafir2023human} in {\bf user-study} of zero-shot human interaction generation. }
        \label{fig:visual-user-study}
        \vspace{-1em}
    \end{figure*}
}

\newcommand{\figControl}{
\begin{figure}[t]
    \centering
    \includegraphics[width=0.6\linewidth]{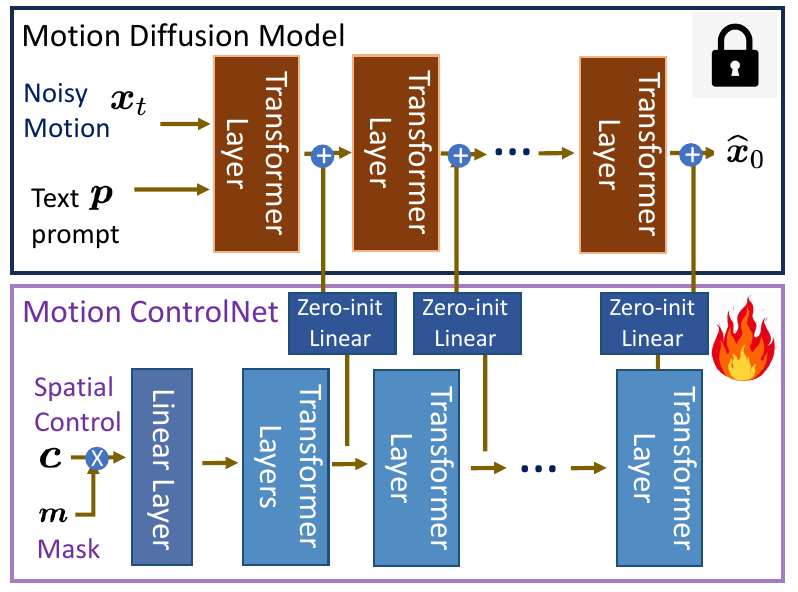}
    \vspace{-1em}
       \caption{\bf Architecture of Motion ControlNet.}
    \label{fig:controlnet}
    
    \vspace{-1em}
\end{figure}
}

%% file: sec/0_abstract.tex
\begin{abstract}
Text-conditioned motion synthesis has made remarkable progress with the emergence of diffusion models. However, the majority of these motion diffusion models are primarily designed for a single character and overlook multi-human interactions. In our approach, we strive to explore this problem by synthesizing human motion with interactions for a group of characters of any size in a zero-shot manner. The key aspect of our approach is the adaptation of human-wise interactions as pairs of human joints that can be either in contact or separated by a desired distance. In contrast to existing methods that necessitate training motion generation models on multi-human motion datasets with a fixed number of characters, our approach inherently possesses the flexibility to model human interactions involving an arbitrary number of individuals, thereby transcending the limitations imposed by the training data. We introduce a novel controllable motion generation method, InterControl, to encourage the synthesized motions maintaining the desired distance between joint pairs. It consists of a motion controller and an inverse kinematics guidance module that realistically and accurately aligns the joints of synthesized characters to the desired location.  Furthermore, we demonstrate that the distance between joint pairs for human-wise interactions can be generated using an off-the-shelf Large Language Model (LLM). Experimental results highlight the capability of our framework to generate interactions with multiple human characters and its potential to work with off-the-shelf physics-based character simulators. Code is available at \url{https://github.com/zhenzhiwang/intercontrol}.
\end{abstract}

%% file: sec/1_intro.tex
\section{Introduction}
\label{sec:intro}

Generating realistic and diverse human motions is a vital task in computer vision, as it has diverse applications in VR/AR, games, and films. In recent years, great progress has been achieved in human motion generation by introducing VAE~\cite{DBLP:journals/corr/KingmaW13}, Diffusion Model~\cite{DBLP:conf/nips/HoJA20,DBLP:conf/iclr/0011SKKEP21} and large language models~\cite{DBLP:conf/nips/BrownMRSKDNSSAA20}. These methods commonly investigated single-person motion generation given texts or action classes~\cite{guo2022tm2t,guo2022t2m,petrovich2022temos,zhang2022motiondiffuse,DBLP:conf/iclr/TevetRGSCB23,DBLP:conf/cvpr/ChenJLHFCY23,guo2020action2motion,petrovich2021action}, part of motion~\cite{duan2021single,harvey2020robust,DBLP:conf/iclr/TevetRGSCB23}, or other related modalities~\cite{li2021ai,li2022danceformer,DBLP:conf/cvpr/TsengCL23,DBLP:journals/tog/AoGLCL22,DBLP:conf/siggraph/HabibieESANNT22}, yet overlooked multi-person interactions. By naively putting their generated single-person motions in a shared global space, such motions could easily penetrate each other. They cannot even perform simple interactions like handshaking due to lack of the ability to control two people's hands to reach the same location at the same time. Many multi-person datasets~\cite{CMU,DBLP:conf/cvpr/GuoBAM22,DBLP:conf/3dim/MehtaSMX0PT18,DBLP:conf/eccv/MarcardHBRP18} lacks text annotations and focus on motion completion given prefix motions. Recently, InterGen~\cite{liang2023intergen} collected a two-person interaction generation dataset, and let model to learn two-person motions from data. It is limited by the fixed number of characters and cannot generalize to arbitrary numbers. Previous methods commonly ignore a good design for general interaction modeling.


This paper investigates a special yet widely used form of human interactions: interactions that could be quantitatively described by spatial relations of human joints, such as distances or orientations, as shown in Fig.~\ref{fig:teaser} (a) and (b). Such interactions are conceptually simple, as their semantics are almost from spatial relations. Thus, they do not require additional interaction data. It only needs pretrained models from single-person data and could be generalized to an arbitrary number of humans. We define {\em human interactions} as steps of {\em joint-joint contact pairs} and devise a single-person motion generation model to take such contact pairs as control signals. Besides, orientations could also be used in control, such as making two people face each other. In this way, interaction generation is transformed to controllable motion generation. Inspired by~\cite{xiao2023unified}, we adapt descriptions of interactions as joint contact pairs by leveraging Large Language Models (LLMs). Thus, human interactions are annotation-free, and interactions could also involve multiple human joints.

As interactions are adapted to our defined joint contact pairs, the key challenge to generate interactions is the {\em precise spatial control} to satisfy the constraint of spatial controls. This difficulty lies in two parts: (1) the discrepancy between {\em control signals in global space} and {\em relative motion representation} in mainstream pretrained models~\cite{guo2022t2m,DBLP:conf/iclr/TevetRGSCB23}: As semantics of motions are independent to global locations, previous works~\cite{guo2022t2m,DBLP:conf/iclr/TevetRGSCB23} commonly utilize the relative motions, where global locations could only be inferred by aggregating velocities. It poses challenges to control local human poses with global conditions. Previous attempts~\cite{DBLP:conf/iclr/TevetRGSCB23,shafir2023human} exploit the inpainting ability of a pretrained model, yet they are unable to control global joints. GMD~\cite{karunratanakul2023guided} proposes a two-stage model of separated root trajectory generation and local pose generation. Although it manages to control root positions, controlling {\em every joint at any time} is still infeasible. (2) the {\em sparse} control signals in the motion sequence: Control signals could be sparse in both temporal and joint dimension, model needs to adaptively adjust trajectories in uncontrolled frames to satisfy the intermittent constraints. 

In this paper, we propose InterControl, a novel human interaction generation method that is able to precisely control the position of any joint at any time for any person, and it is only trained on single-person motion data. By adding spatial controls to MDM~\cite{DBLP:conf/iclr/TevetRGSCB23}, InterControl is a unified framework of two types of spatial control modules: (1) {\em Motion ControlNet} inspired by ControlNet~\cite{zhang2023adding}: It is initialized from a pretrained MDM~\cite{DBLP:conf/iclr/TevetRGSCB23} and takes global spatial locations as input for joint control in the global space. It is able to generate coherent and high-fidelity motions yet joint positions in global space are not perfect. (2) {\em Inverse Kinematics (IK) Guidance} for joint locations: To further align generated motions and spatial conditions precisely, we use inverse kinematics (IK)~\cite{DBLP:conf/cvpr/PavlakosCGBOTB19} to guide the denoising steps towards desired positions. It could be regarded as a classifier guidance~\cite{DBLP:conf/nips/DhariwalN21}, yet it has no extra classifiers. We utilize L-BFGS~\cite{DBLP:journals/mp/LiuN89} as the optimizer to directly align the global conditions in the local space. With two proposed modules, InterControl is able to control multiple joints of any person at any time. Furthermore, InterControl is able to jointly optimize multiple types of spatial controls, such as orientation alignment, collision avoidance, and joint contacts, as long as the distance measures in IK guidance are differentiable. By exploiting its joint control ability, our model is able to generate multi-person interactions with rich contacts, where no multi-person interaction datasets are needed. Our generated interactions could further serve as the reference motion to generate physical animation with meaningful human-wise reactions in simulators. As shown in Fig.~\ref{fig:teaser} (c), one character could actually hit down the other with his fists by taking our generated fighting motions as input. Extensive experiments in HumanML3D~\cite{guo2022t2m} and KIT-ML~\cite{DBLP:journals/bigdata/PlappertMA16} datasets quantitatively validates our joint control ability, and the user study on generated interactions shows a clear preference over previous methods.

To summarize, our contributions are twofold: (1) We are the first to generate multi-person interactions with a single-person motion generation model in a zero-shot manner. (2) We are the first to perform precise spatial control of every joint in every person at any time for interaction generation. 

%% file: sec/2_related.tex
\section{Related Work}
\label{sec:related}

\subsection{Human Motion Generation}

Synthesizing human motions is a long-standing topic. Previous efforts integrate extensive multimodal data as condition to facilitate conditional human motion generation, including text~\cite{guo2022tm2t,guo2022t2m,petrovich2022temos,zhang2022motiondiffuse,DBLP:conf/iclr/TevetRGSCB23,DBLP:conf/cvpr/ChenJLHFCY23,DBLP:conf/aaai/KimKC23}, action label~\cite{guo2020action2motion,petrovich2021action}, part of motion~\cite{duan2021single,harvey2020robust,DBLP:conf/iclr/TevetRGSCB23}, music~\cite{li2021ai,li2022danceformer,DBLP:conf/cvpr/TsengCL23}, speech~\cite{DBLP:journals/tog/AoGLCL22,DBLP:conf/siggraph/HabibieESANNT22} and trajectory~\cite{DBLP:conf/cvpr/Rempe0P0KKFL23,karunratanakul2023guided,DBLP:conf/3dim/KaufmannA0PZH20}. As texts are free-form information that convey rich semantics, recent progress in motion generation are mainly based on text conditions. For example, FLAME~\cite{DBLP:conf/aaai/KimKC23} introduces transformer~\cite{DBLP:conf/nips/VaswaniSPUJGKP17} to process variable-length motion data and language description. MDM~\cite{DBLP:conf/iclr/TevetRGSCB23} introduces the diffusion model and uses classifier-free guidance for text-conditioned motion generation. MLD~\cite{DBLP:conf/cvpr/ChenJLHFCY23} further incorporates a VAE~\cite{DBLP:journals/corr/KingmaW13} to encode motions into vectors and makes the diffusion process in the latent space. Physdiff~\cite{yuan2023physdiff} integrates physical simulators as constraints in the diffusion process to make the generated motion physically plausible and reduce artifacts. PriorMDM~\cite{shafir2023human} treats pretrained MDM~\cite{DBLP:conf/iclr/TevetRGSCB23} as a generative prior and controls MDM by motion inpainting. Our InterControl also use a pretrained MDM, yet we further train a Motion ControlNet instead of using inpainting. A concurrent work OmniControl~\cite{xie2023omnicontrol} also incorporate classifier guidance~\cite{DBLP:conf/nips/DhariwalN21} and controlnet~\cite{zhang2023adding} modules to control all joints in MDM, yet it focuses on single-person motion generation and does not investigate human interaction generation.

\subsection{Human-related Interaction Generation.} As human motions could be affected or interacted by surrounding humans~\cite{DBLP:conf/siggraph/ZhangGYHTW23,DBLP:journals/jvca/KimSK21,DBLP:journals/tvcg/VaillantBK17}, objects~\cite{xu2023interdiff,DBLP:journals/tog/StarkeZKS19,DBLP:journals/cgf/GhoshDGTS23,kulkarni2023nifty,jiang2022chairs} and scenes~\cite{DBLP:conf/cvpr/WangYDL21,DBLP:conf/nips/WangCLZLH22,xiao2023unified,zhao2023synthesizing,DBLP:conf/iccv/HassanCVSYZB21,wang2021synthesizing}, generating interactions is also an important topic. Previous  methods are mainly about human-scene/object interaction. For example, Interdiff~\cite{xu2023interdiff} uses the contact point of human joints and objects as the root to generate object motions. UniHSI~\cite{xiao2023unified} exploits LLM to generate contact steps between human joints and scene parts as an action plan and control the agent perform the plan via reinforcement learning. As previous human-human interactions datasets~\cite{DBLP:conf/3dim/MehtaSMX0PT18,DBLP:conf/eccv/MarcardHBRP18} only contains very few multi-person sequences, previous human-human interaction methods~\cite{DBLP:conf/nips/WangXNW21,DBLP:conf/iclr/0002WG23} are mainly limited to unsupervised motion completion without texts. Recently, InterHuman dataset~\cite{liang2023intergen} is proposed for text-conditioned multi-person interaction generation, yet it only consider the two-person situation and is not able to model more people's interaction. To the best of our knowledge, we are the first to enable a single-person text-conditioned motion generation model to perform interactions between a group of people by controlling diverse joints of each person.

\subsection{Controllable Diffusion Models}
Diffusion-based generative models have achieved great progress in generating various modalities, such as image~\cite{DBLP:conf/cvpr/RombachBLEO22,ho2022classifier,DBLP:conf/nips/DhariwalN21,DBLP:conf/iclr/0011SKKEP21}, video~\cite{esser2023structure,guo2023animatediff,ho2022imagen} and audio~\cite{kong2020diffwave}. Conditions and controlling ability in diffusion models are also well studied: (1) Inpainting-based methods~\cite{DBLP:conf/nips/ChungSRY22,DBLP:conf/iccv/ChoiKJGY21} predict part of the data with the observed parts as condition and rely on diffusion model to generate consistent output, which is used in PriorMDM~\cite{shafir2023human}. (2) Classifier-guidance~\cite{DBLP:conf/nips/DhariwalN21} trains a separate classifier and exploits the gradient of classifier to guide the diffusion process. Our InterControl inherits the spirit of classifier-guidance, yet our guidance is provided by Inverse Kinematics (IK) and no classifier is needed. (3) Classifier-free guidance~\cite{ho2022classifier} trains a conditional and an unconditional diffusion model simultaneously and trade-off its quality and diversity by setting weights. (4) ControlNet~\cite{zhang2023adding} introduces a trainable copy of pretrained diffusion model to process the condition and freezes the original model to avoid degeneration of generation ability. It enables diverse types of dense control signals for various purpose with minimal finetuning effort. Our InterControl also incorporate the idea of ControlNet~\cite{zhang2023adding} to finetune the pretrained MDM~\cite{DBLP:conf/iclr/TevetRGSCB23} to process spatial control signals and improve the quality of generated motions after joint control.

%% file: sec/3_method.tex
\section{InterControl}
\label{sec:method}
InterControl aims to generate interactions with only single-person motion data by precisely controlling every joint of every person at any time, conditioned on text prompts and joint relations. We first formulate interaction generation in Sec.~\ref{sec:interaction_define}, and then introduce control modules for a single-person motion diffusion model in Sec.~\ref{sec:controlnet} and Sec.~\ref{sec:guidance}. Finally we show details to generate interactions from our model in Sec.~\ref{sec:inter}.

\subsection{Formulation of Interaction Generation}
\label{sec:interaction_define}
Inspired by human-scene interaction~\cite{xiao2023unified}, we define human interactions as joint contact pairs $\mathcal{C} = \left\{\mathcal{S}_1, \mathcal{S}_2, \ldots\right\}$, where $\mathcal{S}_i$ is the $i^{th}$ contact step. Taking two-person interaction as an example, each step $\mathcal{S}$ has several contact pairs $\mathcal{S}=\left\{\left\{j^1_1, j^2_1, t^s_1, t^e_1, c_1, d_1\right\},\left\{j^1_2, j^2_2, t^s_2, t^e_2, c_2, d_2\right\}, \ldots\right\}$, where $j^1_k$ is the joint of person 1, $j^2_k$ is the joint of person 2, $t^s_k$ and $t^e_k$ means the start and end frame of the interaction, $c_k$ means contact type from \{contact, avoid\} to pull or push the joint pairs, $d_k$ is the desired distance in the interaction. By converting the contact pairs $\mathcal{S}$ to the mask $\boldsymbol{m}$ and distance $d$, and taking others' joint positions as condition, we could guide the multi-person motion generation process to interact between joints in the form of spatial distance. In this way, interaction generation is transformed to be controllable single-person motion generation taking a text prompt $\boldsymbol{p}$ and a spatial control signal $\boldsymbol{c} \in \mathbb{R}^{N \times J \times 3}$ as input. Its goal is to predict motion sequence $\boldsymbol{x} \in \mathbb{R}^{N \times D}$ whose joints in the global space is aligned with spatial control $\boldsymbol{c}$, where $N$ is number of frames, $J$ is number of joints (e.g., 24 in SMPL~\cite{DBLP:journals/tog/LoperM0PB15}), and $D$ is the dimension of relative joint representations (e.g., 263 in HumanML3D~\cite{guo2022t2m}). Incorporating spatial control in motion generation presents challenges due to the discrepancy between relative motion representation $\boldsymbol{x}$ and global $\boldsymbol{c}$.

\subsection{Human Motion Diffusion Model (MDM)}
\label{sec:mdm}
\noindent{\bf Relative Motion Representation.} HumanML3D~\cite{guo2022t2m} dataset proposes a widely-used~\cite{DBLP:conf/iclr/TevetRGSCB23,yuan2023physdiff,shafir2023human,DBLP:conf/cvpr/ChenJLHFCY23} relative motion representation, and is proved to be easier to learn realistic motions, as the semantics of human motion is independent of global positions. It consists of root joint velocity, other joints' positions, velocities and rotations in the root space, and foot contact labels. To convert it to the global space, root velocities are aggregated, then other joints will be computed based on root. Please refer to Sec.~5 of HumanML3D~\cite{guo2022t2m} for details. 
Due to such discrepancy, previous inpainting-based methods~\cite{DBLP:conf/iclr/TevetRGSCB23,shafir2023human} is not able to control MDM in global space. GMD~\cite{karunratanakul2023guided} decouples motion generation to two separated generation process of root trajectory and pose relative to root, yet it can only control root joint. Directly adopting global joint positions to generate motions yields unnatural human poses, such as unrealistic limb lengths.

\noindent{\bf Diffusion Process in MDM.} Motivated by the success of image diffusion models~\cite{ho2022classifier,DBLP:conf/cvpr/RombachBLEO22,zhang2023adding,DBLP:conf/nips/DhariwalN21,DBLP:conf/iclr/0011SKKEP21}, Motion Diffusion Model (MDM)~\cite{DBLP:conf/iclr/TevetRGSCB23} is proposed to synthesize sequence-level human motions conditioned on texts $\boldsymbol{p}$ via classifier-free guidance~\cite{ho2022classifier}. The diffusion process is modeled as a noising Markov process 
$q\left(\boldsymbol{x}_t \mid \boldsymbol{x}_{t-1}\right)=\mathcal{N}\left(\sqrt{\alpha_t} \boldsymbol{x}_{t-1},\left(1-\alpha_t\right) \boldsymbol{I}\right)$, 
where $\alpha_t \in (0,1)$ are small constant hyper-parameters, thus $\boldsymbol{x}_T \sim \mathcal{N}(0, \boldsymbol{I})$ if $\alpha_t$ is small enough. Here $\boldsymbol{x}_t \in \mathbb{R}^{N \times D}$ is the entire motion sequence at denoising time-step $t$, and there are $T$ time-steps in total. Thus, $\boldsymbol{x}_0$ is the clean motion sequence, and $\boldsymbol{x}_T$ is a random noise to be sampled. The denoising Markov process is defined as 
$p_{\theta}\left(\boldsymbol{x}_{t-1} \mid \boldsymbol{x}_{t}, \boldsymbol{p}\right)=\mathcal{N}\left(\boldsymbol{\mu}_\theta(\boldsymbol{x}_{t}, t, \boldsymbol{p}),\left(1-\alpha_t\right) \boldsymbol{I}\right),$ 
where $\boldsymbol{\mu}_\theta(\boldsymbol{x}_{t}, t, \boldsymbol{p})$ is the estimated posterior mean for the $t-1$ step from a neural network based on the input $\boldsymbol{x}_{t}$ and $\theta$ is its parameters. Following MDM, we predict the clean motion $\boldsymbol{x}_0(\boldsymbol{x}_{t},  t, \boldsymbol{p}; \theta)$ instead of the noise $\boldsymbol{\epsilon}$ via a transformer~\cite{DBLP:conf/nips/VaswaniSPUJGKP17}, and the posterior mean $\boldsymbol{\mu}_\theta(\boldsymbol{x}_{t}, t, \boldsymbol{p})$ is 
\vspace{-5pt}
\begin{equation}
\label{equ:posterior}
\boldsymbol{\mu}_\theta(\boldsymbol{x}_{t}, t,\boldsymbol{p})=\frac{\sqrt{\bar{\alpha}_{t-1}} \beta_t}{1-\bar{\alpha}_t} \boldsymbol{x}_0(\boldsymbol{x}_{t},  t, \boldsymbol{p}; \theta)+\frac{\sqrt{\alpha_t}\left(1-\bar{\alpha}_{t-1}\right)}{1-\bar{\alpha}_t} \boldsymbol{x}_t,
\vspace{-5pt}
\end{equation}
where $\beta_t = 1 - \alpha_t$ and $\bar{\alpha}_t=\prod_{s=0}^t \alpha_s$. MDM's parameter $\theta$ is trained by minimizing the $\ell_2$-loss $\left\|\boldsymbol{x}_0(\boldsymbol{x}_{t},  t, \boldsymbol{p}; \theta)-\boldsymbol{x}_0^*\right\|_2^2$ where $\boldsymbol{x}_0^*$ is the ground-truth motion and $\boldsymbol{x}_0(\boldsymbol{x}_{t},  t, \boldsymbol{p}; \theta)$ is MDM's prediction of $\boldsymbol{x}_0$ at denoising timestep $t$.

\figOVERVIEW
\subsection{Motion ControlNet for MDM}
\label{sec:controlnet}
As MDM is initially conditioned on texts $\boldsymbol{p}$, it requires fine-tuning to accommodate spatial conditions $\boldsymbol{c}$. This is challenging due to the potential sparsity of $\boldsymbol{c}$ across temporal and joint dimensions: (1) Control may be required for only a few joints, necessitating adaptive adjustment of the remaining joints to preserve realistic motion. (2) Control may be desired for only a select few frames, thus the model must interpolate natural human motions for the rest of the sequence.

Inspired by ControlNet~\cite{zhang2023adding}, we introduce Motion ControlNet to generate realistic and high-fidelity motions guided by condition $\boldsymbol{c}$. It is a trainable copy of MDM, while MDM is frozen in our training process. Each transformer encoder layer in ControlNet is connected to its MDM counterpart via a zero-initialized linear layer. This allows InterControl to commence training from a state equivalent to a pretrained MDM, acquiring a residual feature for $\boldsymbol{c}$ in each layer through back-propagation. To process $\boldsymbol{c}$, the uncontrolled joints, frames, and XYZ-dim are masked as $0$. We find that the vanilla $\boldsymbol{c} \in \mathbb{R}^{N \times 3J}$ is effective enough to control the pelvis (root) joint, yet it is still sub-optimal for other joints. Thus, we design a relative condition indicating the distance from the current positions of each joint to $\boldsymbol{c}$. Suppose $R(\cdot)$ is a forward kinematics (FK) to convert relative motion $\boldsymbol{x} \in \mathbb{R}^{N \times D}$ to global space $R(\boldsymbol{x}) \in \mathbb{R}^{N \times J \times 3}$, the relative condition is $\boldsymbol{c}' = \boldsymbol{c}  - R(\boldsymbol{x})$. To provide additional clues, we also use $\boldsymbol{c}'' = \boldsymbol{c} - R(\boldsymbol{x})^{root}$ to represent the distance from the current root to the desired position. We also use the normal of triangles (pelvis, left/right shoulder) $\boldsymbol{n}^{s}$ and (pelvis, left/right hip) $\boldsymbol{n}^{h}$ to represent the current orientation of human. The final condition passed to ControlNet is $\boldsymbol{c}^{final} = (\boldsymbol{c}' || \boldsymbol{c}'' || \boldsymbol{n}^{s} || \boldsymbol{n}^{h})$, where $||$ is concatenation. Please refer to Appendix~\ref{sec:details_controlnet} for more details.

\noindent{\bf Network Training.} Motion ControlNet is the only part that needs finetuning in our framework, while IK guidance is an optimization method in the test time and the LLM in our framework is an off-the-shelf GPT-4~\cite{DBLP:journals/corr/abs-2303-08774}. We adopt the standard ControlNet~\cite{zhang2023adding} training strategy, and the only difference is the data format: we first convert the relative motion to be global locations by FK, and then use random masks that keeps part of global joints to be non-zero as spatial control signals. The training objective is identical to MDM. The spatial conditions are randomly sampled in the temporal or joint dimension. The training data is single-person data only, e.g., HumanML3D~\cite{guo2022t2m}.
\subsection{Inverse Kinematics (IK) Guidance}
\label{sec:guidance}
While Motion ControlNet can adapt joint positions according to sparse conditions, the alignment between predicted poses and global spatial conditions often lacks precision. As Inverse Kinematics (IK) is a classic method for optimizing joint rotations to achieve specific global positions, we employ it to guide the diffusion process towards spatial conditions at test time in a classifier guidance~\cite{DBLP:conf/nips/DhariwalN21} manner, named IK guidance.

\noindent{\bf IK Guidance on general form of losses.} Inspired by classifier guidance~\cite{DBLP:conf/nips/DhariwalN21} and loss-guided diffusion~\cite{DBLP:conf/icml/SongZYM0KCV23}, we employ losses in the global space to steer the denoising process. IK guidance accommodates various forms of distance measurements, enabling both minimization and maximization for flexible control over joint interactions, such as attraction or repulsion. Given the global position $\boldsymbol{c} \in \mathbb{R}^{N \times J \times 3}$, the distance between a joint and condition is $\boldsymbol{d}_{n j} = \left\|\boldsymbol{c}_{n j}-R(\boldsymbol{\mu}_t)_{n j}\right\|_2$, where $\boldsymbol{\mu}_t$ is short for $\boldsymbol{\mu}_\theta(\boldsymbol{x}_{t}, t,\boldsymbol{p})$ mentioned in Sec.~\ref{sec:mdm}, and $R(\cdot)$ is forward kinematics (FK). To allow the interaction of joints with some given distances $d' \in \mathbb{R}^{N \times J \times 3}$, loss of one joint is $\boldsymbol{l}_{n j} = \text{ReLU} \left(\boldsymbol{d}_{n j} - d'_{n j}\right)$ to make the joint and condition be {\bf contacted} within distance $d'_{n j}$; and it is $\boldsymbol{l}_{n j} = \text{ReLU} \left(d'_{n j} - \boldsymbol{d}_{n j} \right)$ to make the joint and condition be {\bf far away}, where ReLU is a function to keep values $\geq 0$ and set values $\leq 0$ to $0$. Finally, with a binary mask $\boldsymbol{m} \in \{0,1\}^{N \times J \times 3}$, the total loss for all joints and frames is
\vspace{-5pt}
\begin{equation}
\label{equ:guidance}
L(\boldsymbol{\mu}_t, \boldsymbol{c})=\frac{\sum_n \sum_j \boldsymbol{m}_{n j} \cdot \boldsymbol{l}_{n j}}{\sum_n \sum_j \boldsymbol{m}_{n j}},
\end{equation}
As $\ell_2$-loss and FK are highly differentiable, we optimize $L(\boldsymbol{\mu}_t, \boldsymbol{c})$ in Equ.~\ref{equ:guidance} w.r.t $\boldsymbol{\mu}_t$ using the second-order optimizer L-BFGS~\cite{DBLP:journals/mp/LiuN89}, which is commonly used in Inverse Kinematics, rather than first-order gradient methods. Classifier guidance~\cite{DBLP:conf/nips/DhariwalN21} utilizes a pre-trained image classifier to direct the diffusion towards a target image class by the gradient $\nabla_{\boldsymbol{x_t}} \log f_\phi\left(y \mid \boldsymbol{x_t}\right)$, where $f_\phi$ is the classifier, $y$ is image class. Unlike this method, we do not rely on a large neural network classifier. L-BFGS has been demonstrated to better align global positions and offer quicker convergence than first-order methods. We update the posterior mean $\boldsymbol{\mu}_t$ using L-BFGS for $k$ iterations at each denoising step, where $k$ is a hyper-parameter. This optimization facilitates both pull and push types of IK guidance, corresponding to two contact types in our interaction model. To maintain consistency in data distribution between training and inference, we also apply IK guidance when training ControlNet. Additionally, employing IK guidance on $\boldsymbol{x}_0$ eliminates the need for training Motion ControlNet, thus enhancing training efficiency. In practice, using L-BFGS on both $\boldsymbol{x}_0$ and $\boldsymbol{\mu}_t$ can yield satisfactory joint and spatial condition alignment. Detailed algorithm for interaction generation is presented in Appendix~\ref{sec:details_algo}. 

As the root position at frame $n$ is derived from cumulative root velocities up to frame $n$ in FK, a single condition at frame $n$ can influence all preceding root positions. This effect also extends to non-root joints, as their global positions are calculated from the root. Consequently, IK guidance can adaptively modify velocities from the start to frame $n$ to meet the condition at frame $n$. Moreover, IK guidance can control any combination of human joints, frames or XYZ-dims, such as controlling the left hand and right foot at a specific frame $n$.


\subsection{Interaction Generation}
\label{sec:inter}
Inverse Kinematics (IK) guidance can optimize various distance measures to facilitate interactions such as avoiding obstacles, preventing collisions, facilitating face-to-face engagements, or enabling joint contacts between individuals. This method allows for intricate interactions among any human joints for an indefinite number of people, despite being trained exclusively on single-person data. As delineated in Section~\ref{sec:interaction_define}, we characterize interactions as pairs of contacting joints. A notable feature of our IK guidance in generating interactions is that both terms of the IK guidance loss function are predicted, allowing for simultaneous optimization within a single process. Specifically, the single-person loss $L_{single}(\boldsymbol{\mu}_t, \boldsymbol{c})$ transforms into $L_{multi}(\boldsymbol{\mu}^a_t, \boldsymbol{\mu}^b_t)$ for interactions, where $a$ and $b$ represent two individuals. The L-BFGS optimizer concurrently optimizes both participants by minimizing $L_{multi}(\boldsymbol{\mu}^a_t, \boldsymbol{\mu}^b_t)$, with $\boldsymbol{\mu}^a_t$ and $\boldsymbol{\mu}^b_t$ being the respective joints engaged in interaction.
Beyond distance measures, our IK guidance can optimize orientation measures as well. For example, one can calculate a person's orientation through the spatial relationship of their joints, like the cross-product of vectors from the left shoulder to the right and from the pelvis to the head. By setting two individuals' unit orientation vectors to $0$, they can face each other or turn away. To ensure they face each other, we can further adjust the relation between one person's orientation vector and the vector from their head to the other's. Such orientation relationships are vital for producing realistic interactions when we only exploit single-person motion generation ability and can be easily expanded to include larger groups.
Another useful strategy in IK guidance is to prevent collision through joint separation pairs, ensuring that the torso joints of two people (such as pelvis, hips, and spines) maintain a certain distance, thereby reducing the likelihood of collisions when other joints are in contact. Besides, we can also regulate the motion region by confining the root joints within the XZ-plane using IK guidance. For the PyTorch-like code illustrating loss functions that enforce joint contacts, separations, or orientation alignment, please refer to Appendix~\ref{sec:details_algo} for details.

In our framework, interaction generation is realized by using joint-joint contact pairs as control signals. These pairs can be manually crafted by users to create desired interactions, akin to utilizing ControlNet~\cite{zhang2023adding} in image generation. However, manually constructing joint contact pairs can be tedious, so we employ an automatic off-the-shelf GPT-4~\cite{DBLP:journals/corr/abs-2303-08774} as a planner. GPT-4 infers text prompts that describe the actions of multiple people, $\boldsymbol{p}^{multi}$, and converts them into single-person prompts, $\boldsymbol{p}$, and contact plans, $\mathcal{C}$, through prompt engineering. The inputs for the LLM Planner include the multi-person sentences $\boldsymbol{p}^{multi}$, background scenario details $\mathcal{B}$, human joint data $\mathcal{J}$, and predefined instructions, rules, and examples. Specifically, $\mathcal{B}$ encompasses the number of individuals, total motion sequence frames, and video playback speed; $\mathcal{J}$ contains names of all joints (for example, the 22 joint names in HumanML3D~\cite{guo2022t2m}); and the rules outline the joint contact pair format and guide the LLM to generate feasible contacts and timesteps. Our method leverages the pre-trained capabilities of GPT-4 to comprehend human joint relationships from interaction descriptions via prompt engineering without any fine-tuning. Thus, the inference process of our model is not related to LLMs, making our comparison  with other methods be fair. Please refer to Appendix~\ref{sec:detail_llm} for details of prompts and contact plans.

%% file: sec/4_experiment.tex
\input{sec/tables}

\section{Experiments}
\noindent{\bf Datasets.} We conduct experiments on HumanML3D~\cite{guo2022t2m} and KIT-ML~\cite{DBLP:journals/bigdata/PlappertMA16} following MDM~\cite{DBLP:conf/iclr/TevetRGSCB23}. HumanML3D contains 14,646 high-quality human motion sequences from AMASS~\cite{mahmood2019amass} and HumanAct12~\cite{guo2020action2motion}, while KIT-ML contains 3,911 motion sequences with more noises. 

\noindent{\bf Evaluation Protocol.} We adopt metrics suggested by {\it Guo et. al.}~\cite{guo2022t2m} to evaluate the quality of alignment between text and motion, which are Frechet Inception Distance ({\bf FID}), {\bf R-Precision}, and {\bf Diversity}. We also report metrics related to spatial controls following GMD~\cite{karunratanakul2023guided} on HumanML3D dataset, which are {\bf Foot skating ratio}, {\bf Trajectory error}, {\bf Location error} and {\bf Average error}. Please refer to Appendix~\ref{sec:details_metrics} or papers~\cite{guo2022t2m,karunratanakul2023guided} for more details.

\noindent Due to the page limit, we put the implementation details and text-to-motion generation in the Appendix~\ref{sec:implementation} and \ref{sec:t2m_experiment}.

\subsection{Single-Person Controllable Motion Generation}
In Tab.~\ref{table:spatial}, we compare InterControl with other spatially controllable methods~\cite{shafir2023human,karunratanakul2023guided, xie2023omnicontrol}. We also include results of MDM~\cite{DBLP:conf/iclr/TevetRGSCB23} to show the controlling metrics~\cite{karunratanakul2023guided} without spatial control.MDM's trajectory can significantly deviate from the intended path in the absence of control signals, with an average error often exceeding 1m. In contrast, inpainting-based control, unaware of global spatial information, results in considerable divergence, as seen with PriorMDM~\cite{shafir2023human}. GMD~\cite{karunratanakul2023guided} decouples this problem and generates root trajectories in the global space, so it achieves better performance in spatial control metrics. However, its limitation to only the root joint constrains its spatial control and interaction capabilities. Our InterControl could achieve very small errors in spatial control metrics for all-joint control thanks to the power of Inverse Kinematics and L-BFGS optimizer. Meanwhile, Motion ControlNet could ensure the motion data is still in the same distribution with the training set by adapting to the posterior mean updated by IK guidance in its training stage, leading to even better FID than previous methods. It is worth noting that we only use a single model to learn the control strategy for all joints, while previous method~\cite{shafir2023human} needs to train separate models and blend them for multiple joints. Our method achieves similar performance with controlling one joint when extending it to control multiple joints (last two rows in Tab.~\ref{table:spatial}). Compared to the recent concurrent work~\cite{xie2023omnicontrol}, we achieve significantly better FID and Traj./Loc. errors than it in both root joint control or random joint control. It~\cite{xie2023omnicontrol} also shows a notable gap between two form of joint controls (0.310 vs. 0.218), while our method is more robust to joint variants (0.178 vs. 0.159) thanks to our special designs of more inputs in Motion ControlNet. Its R-precision and foot-skating ratio are slightly better than ours, we believe the reason is that their 1-st order optimization tolerates more errors when the joint alignment is hard. It is also supported by their worse Traj./Loc. yet better Avg. err., i.e., their method shows more outliers with large errors. However, their design need much more times of optimization compared to ours (e.g., 100 vs. 5) and leads to longer inference time than ours (120s vs. 80s).

\tabSpatial

\subsection{Zero-Shot Multi-Person Interaction Generation}
To validate our model's interaction generation ability, we analyze the spatial control results in interaction scenarios and perform an user study to qualitatively compare our model with PriorMDM~\cite{shafir2023human}. We also introduce an potential application of our interaction generation method for physics animation.

\begin{table}[t]
\centering
\caption{Evaluation on (left) spatial errors and (right) user preference in interactions. }
\begin{minipage}[t]{0.71\textwidth}
\tabInteraction
\end{minipage}
\hfill
\begin{minipage}[t]{0.28\textwidth}
\tabUser
\end{minipage}
\label{table:inter}
\vspace{-1em}
\end{table}
\noindent{\bf Spatial Control.} In Tab.~\ref{table:inter} (left), we compare spatial-related metrics with PriorMDM in zero-shot human interaction generation. Specifically, we collect $100$ descriptions of two-person actions from InterHuman Dataset~\cite{liang2023intergen} and let an off-the-shelf GPT-4~\cite{DBLP:journals/corr/abs-2303-08774} to adapt them to single-person motion descriptions and joint-joint contact pairs via prompt engineering (see Tab.~\ref{tab:detailed_prompt_example} in Appendix). Then, we utilize an InterControl model pretrained on the HumanML3D dataset to generate human interactions conditioned on text prompts and joint contact pairs. The spatial-related metrics are reported over controlled joints and frames. InterControl achieves good performance of spatial errors in interaction scenarios, indicating its robustness in precise spatial control for multiple humans. In contrast, PriorMDM~\cite{shafir2023human} could only take interaction descriptions as input and unable to perform spatial control, leading to much larger spatial errors.

\noindent{\bf User Study.} We conduct a user study to qualitatively compare our method with PriorMDM on the text-conditioned two-person interaction generation. 134 unique users were participating in the user study, where each user will answer 19 single choice questions to compare our results with PriorMDM~\cite{shafir2023human}. Results in Tab.~\ref{table:inter} (right) shows that our generated interactions are clearly preferred over PriorMDM by a percent of 81.2\%. We also shows an example sequence of qualitative comparison with PriorMDM~\cite{shafir2023human} in the user study in Fig.~\ref{fig:visual-user-study}. PriorMDM~\cite{shafir2023human} shows severe torso collision between two human skeletons and the generated two-people motion is not aligned with the interaction description, while our model has no torso collision thanks to the collision avoidance loss in our IK guidance. Besides, our method also produces reasonable kicking actions between two people according to the semantics of interaction description. Please refer to Appendix~\ref{sec:user_study} for details.
\figUserStudy
\figVisual
 \noindent{\bf Qualitative results:} Although our model is only trained on single-person data, it is still possible to generate interactions between an arbitrary number of people via our designed format of interaction. In Fig.~\ref{fig:visual-skeleton}, we show two representative results of zero-shot interaction generation. (1) Two-person dancing:  In addition to the single person dancing from the pretrained ability of single-person model, we further let them hold hands from time to time and prevent them from collision between their torsos. To further make their dance natural, we also employ a loss to promote their orientations to be face-to-face. (2) Three-person fighting: In addition to a single person performing punching and kicking, we further let them punch or kick others' head and torso, and also prevent their torsos from collision. Compared to existing interaction generation method~\cite{liang2023intergen}, our method is able to generate interaction between any number of people, while InterGen~\cite{liang2023intergen} is only able to generate two-person interaction. Besides, our method is the first method to leverage single-person motion generation model to generate human interactions in a zero-shot manner.

\noindent{\bf Application:} Our method is able to seamlessly integrate with off-the-shelf character simulation approaches, allowing us to synthesize physically plausible human reactions. As shown in Fig.~\ref{fig:teaser} (c), our method synthesizes the motions, where the orange character is fighting with other two characters, as the reference of the SoTA physics-aware motion imitator~\cite{DBLP:conf/iccv/0002CWKX23}. The interactions of our motions are designed to hit heads of other characters with fists. Leveraging the precise spatial control provided by our approach, the animated characters in the simulator can accurately respond to these impacts, resulting in realistic reactions such as being knocked down. This capability to generate spatially coherent multi-human interactions enables our method to improve the plausibility and responsiveness of synthesized reactions within physics-based character animations.

\tabAblation
\tabTime
\subsection{Ablation Studies}
\label{sec:ablation}
To further investigate the effectiveness of InterControl, we ablate our method in Tab.~\ref{table:ablation} and reveal some key information in controlling the motion generation model in the global space. Then we also analyze the computational costs of our method to ensure our control is efficient. We will refer to the variants of InterControl by row numbers in Tab.~\ref{table:ablation}. All experiments are trained on all joints and evaluated with randomly selected joints to report average performance.

\noindent{\bf Motion ControlNet.} By dropping ControlNet, we find that IK guidance could still follow spatial controls with very low errors, yet the motion quality (e.g., FID) is significantly damaged (row 1 vs. row 2). Our ControlNet could adapt to the posterior distribution updated by IK guidance, and produce high-quality motion data. We also find that our $\boldsymbol{c}^{final}$ provides key information in controlling all joints: For root control only, the FID of $\boldsymbol{c}^{final}$ and $\boldsymbol{c}$ shows small difference. However, the FID of root control is always slightly better than all-joint control ($\sim 0.07$) when we use $\boldsymbol{c}$, indicating insufficient information in all-joint control. We alleviate this by introducing extra information in $\boldsymbol{c}^{final}$ for Motion ControlNet and improve the FID of all-joint control from 0.227 (row 3) to 0.178 (row 1).

\noindent{\bf IK guidance.} By dropping IK guidance, Motion ControlNet can produce good semantic-level metrics (e.g., FID) compared with MDM by using extra spatial cues (row 4). However, this variant will lead to more spatial errors and cannot strictly follow spatial controls in global space. As precise joint alignment is vital for interactions, IK guidance is important for our InterControl. Another variant is updating IK guidance on ControlNet's prediction $\boldsymbol{x}_0$ (row 5), instead of the posterior mean $\boldsymbol{\mu}_t$. Its advantage is faster training speed because IK guidance is no longer needed in training ControlNet (similar to classifier guidance~\cite{DBLP:conf/nips/DhariwalN21}) yet it leads to slightly worse FID than using $\boldsymbol{\mu}_t$. We believe the reason is that IK guidance still changes the data distribution in denoising steps even if it is updated on $\boldsymbol{x}_0$. Finally, we also report the result of 1-st order gradient in classifier guidance~\cite{DBLP:conf/nips/DhariwalN21} (row 6) instead of L-BFGS. We find it takes more computations to achieve similar performance with L-BFGS, which is analyzed below.

\noindent{\bf Inference time analysis.} In practice, we find that IK guidance in last few denoising steps (e.g., $t \in [0,9]$) is vital for precise joint control, while most denoising steps $t \in [10,999]$ are less important yet take most of computations. IK guidance on $\boldsymbol{x}_0$ with only once L-BFGS in $t \in [10,999]$ and $10$ times in $t \in [0,9]$ could leads FID $0.234$ in controlling all joints, yet leads to minimal extra computations. We report its total inference time of $1000$ denoising steps by adding sub-modules step-by-step in Tab.~\ref{table:time}. GMD~\cite{karunratanakul2023guided} needs $110$s to run two-stage diffusion models, while we only needs $80$s. Gradient-based optimization in the recent work~\cite{xie2023omnicontrol} needs $120$s to achieve similar control quality. Leveraging GPU parallel computing capabilities, InterControl can efficiently generate motions for a batch of $32$ people in $91$ seconds, enabling efficient group motion generation.

\noindent{\bf Sparse control signals in temporal.} As a key challenge of spatial control is the sparsity, we also report results with sparsely selected frames as control (sparsity = $0.25$ and $0.025$) in Tab.~\ref{table:ablation} (row 7 and 8). Our model demonstrates consistent performance in both spatial error and semantic-level metrics when using sparse signals, e.g., FID $0.255$ and avg. err. $0.0467$ with sparsity $0.025$, while GMD~\cite{karunratanakul2023guided} achieves FID $0.523$ and avg. err. $0.139$ with the same sparsity. 

%% file: sec/tables.tex
\newcommand{\tabHumanml}{
\resizebox{0.99\textwidth}{!}{
\centering
\begin{tabular}{lccc}
\toprule
HumanML3D & FID $\downarrow$~ & \multicolumn{1}{p{1.9cm}}{\centering R-precision $\uparrow$ \\ (Top-3)} & Diversity $\rightarrow$\\ 
 \midrule
 Real & 0.002 & 0.797 & 9.503 \\ 
 \midrule
JL2P~\cite{ahuja2019language2pose} & 11.02 & 0.486 & 7.676  \\ 
Text2Gesture~\cite{bhattacharya2021text2gestures} & 7.664 & 0.345 & 6.409  \\ 
T2M~\cite{guo2022t2m} & 1.067 & 0.740 & 9.188  \\ 
MotionDiffuse~\cite{zhang2022motiondiffuse} & 0.630 & \textbf{0.782} & 9.410 \\
MLD~\cite{DBLP:conf/cvpr/ChenJLHFCY23} & 0.473  & 0.772 & 9.724 \\ 
PhysDiff~\cite{yuan2023physdiff} & 0.433  & 0.631 & - \\  
T2M-GPT~\cite{DBLP:conf/cvpr/ZhangZCZZLSY23} & \textbf{0.116} & 0.775 & 9.761 \\
MotionGPT ~\cite{jiang2023motiongpt} & 0.232 & \underline{0.778} & \underline{9.528} \\ 
MDM~\cite{DBLP:conf/iclr/TevetRGSCB23} & 0.544 & 0.611 & 9.446  \\ 
\midrule
PriorMDM~\cite{shafir2023human} & 0.540 & 0.640 & 9.160 \\ 
GMD~\cite{karunratanakul2023guided} & 0.212 & 0.670 & 9.440 \\ 
OmniControl~\cite{xie2023omnicontrol}  & 0.218   &0.687   &   9.422 \\ 
Our InterControl  & \underline{0.159} & 0.671 & \textbf{9.482} \\ 
\bottomrule
\end{tabular}}
\label{table:text}
}

\newcommand{\tabKit}{
\resizebox{0.99\textwidth}{!}{
\centering
\begin{tabular}{lccc}
\toprule
KIT-ML & FID $\downarrow$~ & \multicolumn{1}{p{1.9cm}}{\centering R-precision $\uparrow$ \\ (Top-3)} & Diversity $\rightarrow$\\ 
 \midrule
 Real & 0.031 & 0.779 & 11.08 \\ 
 \midrule
T2M~\cite{guo2022t2m} & 3.022 & 0.681 & 10.72  \\ 
MotionDiffuse~\cite{zhang2022motiondiffuse} & 1.954 & \underline{0.739} & \textbf{11.10}  \\
MLD~\cite{DBLP:conf/cvpr/ChenJLHFCY23} & \textbf{0.404}  & 0.734 & 10.80  \\  
T2M-GPT~\cite{DBLP:conf/cvpr/ZhangZCZZLSY23} & 0.514 & \textbf{0.745} & \underline{10.92} \\
MotionGPT~\cite{jiang2023motiongpt} & 0.510 & 0.680 & 10.35 \\
MDM~\cite{DBLP:conf/iclr/TevetRGSCB23} & \underline{0.497} & 0.396 & 10.84  \\
\midrule
PriorMDM$^\dagger$~\cite{shafir2023human} & 0.830 & 0.397 & 10.54 \\ 
GMD$^\dagger$~\cite{karunratanakul2023guided} & 1.537 & 0.385 & 9.78 \\
OmniControl~\cite{xie2023omnicontrol} &0.702 &  0.397  &  10.93  \\
Our InterControl  & 0.580 & 0.397 & 10.88 \\ 
\bottomrule
\end{tabular}}

\label{table:text_kit}
}

\newcommand{\tabSpatial}{
\begin{table*}[t]
\centering
\caption{
{\bf Spatial control} results on HumanML3D~\cite{guo2022t2m}. $\rightarrow$ means closer to real data is better.
\textit{Random One/Two/Three} reports the average performance over 1/2/3 randomly selected joints in evaluation. $^\dagger$ means our evaluation on their model.}
\resizebox{1.0\textwidth}{!}{
\begin{tabular}{c|c|ccccccc}
\toprule
 \multirow{2}{*}{Method}  & \multirow{2}{*}{Joint} & \multirow{2}{*}{FID $\downarrow$} & \multicolumn{1}{p{1.9cm}}{\centering R-precision $\uparrow$ \\ (Top-3)} & \multirow{2}{*}{Diversity $\rightarrow$} &   \multicolumn{1}{p{1.8cm}}{\centering Foot skating \\ ratio $\downarrow$ } & \multicolumn{1}{p{1.6cm}}{\centering Traj. err. $\downarrow$ \\ (50 cm)} & \multicolumn{1}{p{1.5cm}}{\centering Loc. err. $\downarrow$ \\ (50 cm)} & \multicolumn{1}{p{1.5cm}}{\centering Avg. err.$\downarrow$ \\ (m)} \\ \midrule
 Real data& - & 0.002  & 0.797 & 9.503 & 0.0000 & 0.0000 & 0.0000 & 0.0000 \\ 
 MDM~\cite{DBLP:conf/iclr/TevetRGSCB23}  & No Control & 0.544  & 0.611 & 9.446 & 0.0943 & 0.8909 &  0.6015 & 1.1843\\ 
 \midrule
 PriorMDM~\cite{shafir2023human}$^\dagger$  &\multirow{4}{*}{Root} & 0.498  & 0.586 & 9.167 &  0.0924 & 0.3726 & 0.2210 &  0.4552\\ 
 GMD~\cite{karunratanakul2023guided}$^\dagger$      & & 0.276  & 0.655 & 9.245 & 0.1108 & 0.0987 & 0.0356& 0.1457 \\ 
 OmniControl~\cite{xie2023omnicontrol} & &0.218 &0.687& 9.422& \textbf{0.0547} &0.0387& 0.0096& \textbf{0.0338}\\
 Ours  & & \textbf{0.159} & 0.671 & 9.482& 0.0729  & \textbf{0.0132} &   \textbf{0.0004}&   0.0496\\ 
 \midrule
  OmniControl~\cite{xie2023omnicontrol} & \multirow{2}{*}{Random one}&0.310& \textbf{0.693} &\textbf{9.502} &0.0608 &0.0617 &0.0107 &0.0404\\
 Ours  &  & 0.178  & 0.669 & 9.498 & 0.0968 & 0.0403 & 0.0031 & 0.0741 \\
 \midrule
Ours  & Random two &0.184 &0.670&9.410&0.0948&0.0475&0.0030&0.0911 \\
Ours  & Random three&0.199 &0.673& 9.352&0.0930&0.0487& 0.0026&0.0969\\
\bottomrule
\end{tabular}
}
\label{table:spatial}
\end{table*}}

\newcommand{\tabAblation}{
\begin{table*}[t]
\centering
\caption{
{\bf Ablation studies} on the HumanML3D~\cite{guo2022t2m} dataset.
}
\resizebox{1.0\textwidth}{!}{
\begin{tabular}{c|c|ccccccc}
\toprule
 \multirow{2}{*}{Item}  & \multirow{2}{*}{Method}& \multirow{2}{*}{FID $\downarrow$} & \multicolumn{1}{p{1.9cm}}{\centering R-precision $\uparrow$ \\ (Top-3)} & \multirow{2}{*}{Diversity $\rightarrow$} &   \multicolumn{1}{p{1.8cm}}{\centering Foot skating \\ ratio $\downarrow$ } & \multicolumn{1}{p{1.6cm}}{\centering Traj. err. $\downarrow$ \\ (50 cm)} & \multicolumn{1}{p{1.5cm}}{\centering Loc. err. $\downarrow$ \\ (50 cm)} & \multicolumn{1}{p{1.5cm}}{\centering Avg. err.$\downarrow$ \\ (m)} \\ \midrule
 (1) &Ours (random joint)  & \textbf{0.178}  & 0.669 & \textbf{9.498} & 0.0968 & 0.0403 & 0.0031 & 0.0741 \\
 \midrule
(2) &w/o ControlNet&  0.965 & 0.621 & 9.216& 0.1624 & 0.0879 & 0.0059 & 0.1013\\
 (3) & w/ original $\boldsymbol{c}$ & 0.227  & 0.656 & 9.544 & 0.1004 & 0.0697 & 0.0042 & 0.0785\\
 \midrule
  (4) & w/o IK guidance& 0.187  & 0.664 & 9.598& \textbf{0.0704}  &  0.8569&  0.4553 & 0.6557\\ 
 (5) &IK guidance on $\boldsymbol{x}_0$&  0.211  &  0.668& 9.394 & 0.1164& 0.0907 & 0.0088& 0.0981\\
 (6) &w/ 1-st order grad&  0.198& 0.668 &9.472 & 0.0987 &0.0879  &0.0096  & 0.0877\\
\midrule
(7) &sparsity = 0.25& 0.248  & \textbf{0.671} & 9.442& 0.0801 & 0.0106 &  0.0007 & 0.0546 \\
(8) &sparsity = 0.025 & 0.255  & 0.663 & 9.520& 0.0705 & \textbf{0.0015} & \textbf{0.0001} & \textbf{0.0067}\\
\bottomrule
\end{tabular}
}

\label{table:ablation}
\end{table*}
}

\newcommand{\tabTime}{

\begin{table}[t]
\centering
\caption{
\textbf{Inference time analysis} on a NVIDIA A100 GPU. 
}
\resizebox{0.9\textwidth}{!}{
\begin{tabular}{c|c|c|c|c}
\toprule
   Sub-Modules & MDM &  + Control Module&   + Guidance $t \in [10,999]$ & + Guidance  $t \in [0,9]$ \\ 
 \midrule
 Time (s)  & 39.1 & 57.3  & 76.5 & 80.1 \\ 
\bottomrule
\end{tabular}
}
\label{table:time}
\vspace{-1em}
\end{table}

}

\newcommand{\tabInteraction}{
\centering
\resizebox{0.9\textwidth}{!}{
\begin{tabular}{c|c|c|c}
\toprule
  Spatial Errors &Traj. err. (20 cm) $\downarrow$ &  Loc. err. (20 cm) $\downarrow$ &   Avg. err. (m) $\downarrow$\\ 
 \midrule
  PriorMDM~\cite{shafir2023human}& 0.6931 & 0.3487  & 0.6723  \\ 
  Ours& {\bf 0.0082} & {\bf 0.0005}  & {\bf 0.0084}  \\ 
\bottomrule
\end{tabular}
}
}

\newcommand{\tabUser}{
\centering
\resizebox{0.9\textwidth}{!}{
\begin{tabular}{c|c}
\toprule
   User-study &Preference\\ 
 \midrule
  PriorMDM~\cite{shafir2023human}& 18.8\%  \\ 
  Ours& {\bf 81.2\%}  \\ 
\bottomrule
\end{tabular}
}
}

%% file: sec/5_conclusion.tex
\section{Conclusion and Limitations}
We presented InterControl, a multi-person interaction generation method that is only trained on single-person motion data. It could generate interactive human motions of an arbitrary number of people. We achieve this by enabling a text-conditioned motion generation model with the ability to control every joint of every person at any time. We propose two complementary modules, named Motion ControlNet and IK guidance, to improve both the spatial alignment between joints and desired positions, and the overall quality of whole motions. Extensive experiments are conducted on HumanML3D and KIT-ML benchmarks to validate the effectiveness and efficiency of our proposed modules. We enable InterControl the ability of text-conditioned interaction generation by leveraging the knowledge of LLMs. Qualitative results and user study validate that InterControl could generate high-quality interactions by precise spatial joint control.

\noindent{\bf Limitations.} As InterControl is not trained on multi-person data, its definition of interaction is based on distances (being {\em contacted} or {\em separated}) or orientations. Its motion quality is from motion generation model trained on single-person motion data, and the plausibility of interactions is from the knowledge of LLMs, i.e., to what extent the joint contact pairs are consistent to the semantics of interaction descriptions. Yet, InterControl could generate interactions of an arbitrary number of people, while all existing interaction generation methods cannot.

%% file: sec/supp.tex
\section{More Details about InterControl}
\label{sec:detail}

\begin{algorithm}[h]
\small
\caption{Two-people interaction model inference}
\label{alg:inter}
\begin{algorithmic}[1]

\Require a Motion Diffusion Model $M$ with parameter $\theta$, a Motion ControlNet $C$ with parameter $\phi$, interaction prompts $\boldsymbol{p}^{multi}$, number of L-BFGS $K$, Forward Kinematics operation FK, masked selection operation $S$.
\State $\boldsymbol{x}^a_T, \boldsymbol{x}^b_T \sim \mathcal{N}(0, \boldsymbol{I})$
\For{$t$ from $T$ to $1$}
  \State \text{\# LLM-Planner} 
  \State $\boldsymbol{p}^a, \boldsymbol{p}^b, \text{mask} \leftarrow \text{LLM} (\boldsymbol{p}^{multi})$
  \State \text{\# Copy Spatial Condition from Each Other} 
  \State $\boldsymbol{c}^a \leftarrow S (\text{FK} (\boldsymbol{x}^b_t), \text{mask})$
  \State $\boldsymbol{c}^b \leftarrow S (\text{FK} (\boldsymbol{x}^a_t), \text{mask})$
  \State \text{\# Motion ControlNet}
    \State $\{\boldsymbol{f}\}^a \leftarrow C\left(\boldsymbol{x}^a_t, t, \boldsymbol{p}^a, \boldsymbol{c}^{a}; \phi\right)$
    \State $\{\boldsymbol{f}\}^b \leftarrow C\left(\boldsymbol{x}^b_t, t, \boldsymbol{p}^b, \boldsymbol{c}^{b}; \phi\right)$
    \State \text{\# Motion Diffusion Model}
    \State $\boldsymbol{x}^a_0 \leftarrow M\left(\boldsymbol{x}^a_t, t, \boldsymbol{p}^a,\{\boldsymbol{f}\}^a ; \theta\right)$
    \State $\boldsymbol{x}^b_0 \leftarrow M\left(\boldsymbol{x}^b_t, t, \boldsymbol{p}^b,\{\boldsymbol{f}\}^b ; \theta\right)$
    \State $\boldsymbol{\mu}^a_t, \Sigma_t \leftarrow \mu\left(\boldsymbol{x}^a_0, \boldsymbol{x}^a_t \right), \Sigma_t$ \text{\quad \# Posterior}
    \State $\boldsymbol{\mu}^b_t, \Sigma_t \leftarrow \mu\left(\boldsymbol{x}^b_0, \boldsymbol{x}^b_t \right), \Sigma_t$ \text{\quad \# Posterior}
    \For{$k$ from $1$ to $K$}
    \State \text{\# IK guidance}
        \State $\boldsymbol{\mu}^a_t, \boldsymbol{\mu}^b_t \leftarrow \text{L-BFGS} (L(\boldsymbol{\mu}^a_t, \boldsymbol{\mu}^b_t))$ 
    \EndFor
    \State $\boldsymbol{x}^a_{t-1} \sim \mathcal{N}(\boldsymbol{\mu}^a_t, \Sigma_t)$
    \State $\boldsymbol{x}^b_{t-1} \sim \mathcal{N}(\boldsymbol{\mu}^b_t, \Sigma_t)$
\EndFor
\\
\Return $\boldsymbol{x}^a_0, \boldsymbol{x}^b_0$
\end{algorithmic}
\end{algorithm}

\subsection{Pseudo-code of IK guidance}
\label{sec:details_algo}
Here we elaborate the details of IK guidance's algorithm. As we mentioned in the main paper, IK guidance could be performed on predicted clean motion (i.e., $\boldsymbol{x}_0$) or posterior mean in denoising step $t$ (i.e., $\boldsymbol{\mu}_t$). In practice, we find that $\boldsymbol{x}_0$ works well in root control, and it does not require IK guidance in training Motion ControlNet, leading to faster training speed. Besides, it also requires less times of L-BFGS~\cite{DBLP:journals/mp/LiuN89}, which means faster inference speed. $\boldsymbol{\mu}_t$ leads to better FID in controlling all joints, yet it requires more times of L-BFGS~\cite{DBLP:journals/mp/LiuN89} and also need IK guidance in training Motion ControlNet. We show the pseudo-code of InterControl in interaction generation in Algorithm~\ref{alg:inter}. 

\subsection{Details of Motion ControlNet}
\label{sec:details_controlnet}
In this subsection, we elaborate the details of Motion ControlNet's architecture.
Motion ControlNet is designed to adaptively generate realistic and high-fidelity motion sequences based on condition $\boldsymbol{c}$. It is a trainable copy of MDM, and each transformer encoder layer of ControlNet and the original MDM is connected by a zero-initialized linear layer, as shown in Fig.~\ref{fig:controlnet}. The parameters in the original MDM is pretrained and frozen in the entire training process. Thus, our framework in the finetuning process starts from the weights that is equivalent to a pretrained MDM due to the zero-initialized linear layers. ControlNet will learn a residual feature for spatial control signals $\boldsymbol{c}$ in each transformer layer by the back-propagated gradients. Thus, our model is able to implicitly adjust model weights for all joints and frames based on a sparse spatial condition $\boldsymbol{c}$ by learning the spatial-level conditional distribution in addition to the semantic-level distribution. 

To process condition $\boldsymbol{c}$, the uncontrolled joints, frames and XYZ-dim are masked as $0$. Then we use a linear layer to project the condition $\boldsymbol{c} \in \mathbb{R}^{N \times 3J}$ to the hidden dimension of transformer layers as $\boldsymbol{c}^H \in \mathbb{R}^{N \times D^H}$, and feed $\boldsymbol{c}'$ to transformer encoder layers in ControlNet. We use a zero-initialized linear layer to link the output of each layer in ControlNet to the transformer encoder layer of pretrained and frozen MDM via a residual connection~\cite{DBLP:conf/cvpr/HeZRS16}. We use extra information as condition for Motion ControlNet $\boldsymbol{c}^{final} = cat (\boldsymbol{c}', \boldsymbol{c}'', \boldsymbol{n}^{s}, \boldsymbol{n}^{h})$. The details of $\boldsymbol{c}^{final}$ has been explained in Sec 3.3 in our main paper.

\figControl
\subsection{LLM-Planner}
\label{sec:detail_llm}
In this section, we further elaborate the details of LLM Planner. Specifically, we collect $100$ sentences describing human interactions with joint contacts from the description of InterHuman Dataset~\cite{liang2023intergen}. Then, we use a GPT-4~\cite{DBLP:journals/corr/abs-2303-08774} with the prompt in Tab.~\ref{tab:detailed_prompt_example} to let GPT-4 to produce joint-joint contact plans for us. For each collected sentence, we replace it as the {\em instruction} in the prompt, and LLM will generate $10$ task plans for us, as shown in Tab.~\ref{tab:task_plan}. We manually correct typos of task plans generated by LLM, such as typos of joint name, invalid joint name, or invalid start frame or end frame. It leads to $989$ valid task plans. Finally, we write Python scripts to transform the natural language tasks plans to Python Json format, as shown in Tab.~\ref{tab:json}. We take single-person language prompts in task plans as texts for motion diffusion model, and transform information in 'steps' to joint contact masks in the spatial condition. Specifically, we update the other person's joint positions as the current person's spatial condition in each denoising step, and use the spatial condition to guide Motion ControlNet and IK guidance in the same way with single-person scenarios. We evaluate the quality of interactions by using metrics like trajectory error and average error proposed by GMD~\cite{karunratanakul2023guided} in the same way with single-person scenarios. We only evaluate on joints and frames in the joint-joint contact pairs. The result on our collected $989$ task plans is shown in Tab. 5 in the main paper.

\section{Additional Experiments}
\label{sec:more_experiment}

\subsection{Implementation Details.}
\label{sec:implementation}
We initialize parameters of both original MDM and Motion ControlNet from pretrained MDM~\cite{DBLP:conf/iclr/TevetRGSCB23} weight and freeze the parameters of original MDM during training. Following MDM~\cite{DBLP:conf/iclr/TevetRGSCB23}, we use CLIP~\cite{DBLP:conf/icml/RadfordKHRGASAM21} model to encode text prompts. We run L-BFGS~\cite{DBLP:journals/mp/LiuN89} in IK guidance $5$ times for the first $990$ denoising steps and $10$ times for the last $10$ denoising steps on the posterior mean $\boldsymbol{\mu}_t$; and once for the first $990$ steps and $10$ times for the last $10$ steps on clean motion $\boldsymbol{x}_0$. We use IK guidance in training ControlNet when using it on $\boldsymbol{\mu}_t$. We set two types of mask $\boldsymbol{m} \in \{0,1\}^{N \times J \times 3}$: (1) Only keeps pelvis (root) joint for root control to fairly compare with previous methods; (2) Randomly keep one joint in each iteration to learn to control all joints for interaction generation. Each type of mask will be used in both training and inference for consistency. Thus, we get two model weights, where (1) could be fairly compared with previous methods and we use (2) for interaction generation. We use AdamW~\cite{DBLP:conf/iclr/LoshchilovH19} optimizer and set the learning rate as 1e-5.

\begin{table}[t]
\centering
\caption{{\bf Text-to-motion evaluation} on the (left) HumanML3D~\cite{guo2022t2m} and (right) KIT-ML~\cite{DBLP:journals/bigdata/PlappertMA16} datasets. The right arrow $\rightarrow$ means closer to real data is better. Methods in the upper part are unable to perform spatial control. $^\dagger$ means our implementation.}
\begin{minipage}[t]{0.48\textwidth}
\tabHumanml
\end{minipage}
\hfill
\begin{minipage}[t]{0.51\textwidth}
\tabKit
\end{minipage}
\vspace{-1em}
\end{table}

\subsection{Text-to-Motion Generation Results}
\label{sec:t2m_experiment}
To generally compare our InterControl with previous text-conditioned motion generation methods, we report the alignment quality of text and generated motions suggested by {\it Guo et. al.}~\cite{guo2022t2m} in Tab.~\ref{table:text}. Note that methods in the upper part of both tables are unable to perform spatial control, thus they are incapable of generating controllable motions and interactions even if they have lower FID or higher R-precision. For instance, T2M-GPT~\cite{DBLP:conf/cvpr/ZhangZCZZLSY23} and MotionGPT~\cite{jiang2023motiongpt} tokenize human poses to discrete tokens and is unable to incorporate any spatial information. MLD~\cite{DBLP:conf/cvpr/ChenJLHFCY23} uses latent diffusion to accelerate denoising steps, yet performing spatial control needs to convert each step of latent feature back to motion representations. It leads to much more computation than MDM~\cite{DBLP:conf/iclr/TevetRGSCB23} and is opposite to MLD's motivation of latent diffusion. Among methods that are suitable for spatial control~\cite{shafir2023human,karunratanakul2023guided} in Tab.~\ref{table:text}, InterControl achieves the best performance in most of semantic-level metrics, and is better than the recent work OmniControl~\cite{xie2023omnicontrol} that focuses on single-person motion yet shares similar design of spatial controlling with us.

\begin{table*}[t]
\centering
\resizebox{0.99\textwidth}{!}{
\begin{tabular}{c|c|ccccccc}
\toprule
 \multirow{2}{*}{Method}  & \multirow{2}{*}{Joint} & \multirow{2}{*}{FID $\downarrow$} & \multicolumn{1}{p{1.9cm}}{\centering R-precision $\uparrow$ \\ (Top-3)} & \multirow{2}{*}{Diversity $\rightarrow$} &   \multicolumn{1}{p{1.8cm}}{\centering Foot skating \\ ratio $\downarrow$ } & \multicolumn{1}{p{1.6cm}}{\centering Traj. err. $\downarrow$ \\ (50 cm)} & \multicolumn{1}{p{1.5cm}}{\centering Loc. err. $\downarrow$ \\ (50 cm)} & \multicolumn{1}{p{1.5cm}}{\centering Avg. err.$\downarrow$ \\ (m)} \\ \midrule
 Ours (all)  & Root & 0.184 &0.672  &9.315 & 0.1044 & 0.0317& 0.0018 & 0.0693 \\ 
 Ours (all)  & Left foot & 0.242  & 0.664 & 9.184 & 0.1005 & 0.0696 & 0.0024 & 0.0671 \\ 
 Ours (all)  & Right foot & 0.236  & 0.669 & 9.201 & 0.0983 & 0.0798 & 0.0029 & 0.0680 \\ 
 Ours (all)  & Head & 0.172 & 0.678 & 9.359 & 0.0958 & 0.0523& 0.0044 & 0.0846 \\ 
 Ours (all)  & Left wrist & 0.260 &  0.660& 8.965& 0.0915 & 0.0375& 0.0012 &  0.0874\\ 
 Ours (all)  & Right wrist & 0.284 & 0.655 & 9.003 & 0.0920 & 0.0364& 0.0010 & 0.0872 \\ 
 \bottomrule
\end{tabular}
}
\caption{
{\bf Spatial control} results on the HumanML3D~\cite{guo2022t2m} dataset.
\textit{Ours (all)} means the model is trained on one randomly selected joint among all joints in each iteration.
}
\label{table:joint_spatial}
\end{table*}

\subsection{More Single-joint Control Results}
\label{sec:more_joint_control}
In Tab. 1 of our main paper, we have shown the spatial control results with root joint and randomly selected one/two/three joints. Following the recent work~\cite{xie2023omnicontrol}, we also show the spatial control performance on specific joints in Tab.~\ref{table:joint_spatial}. We find that feet and hands are more difficult to control due to their flexibility, while root (pelvis) and head are more easier to follow, leading to better FID and R-precision.

\subsection{Details of User Study}
\label{sec:user_study}
\begin{figure}[t]
    \centering
    \includegraphics[width=0.8\linewidth]{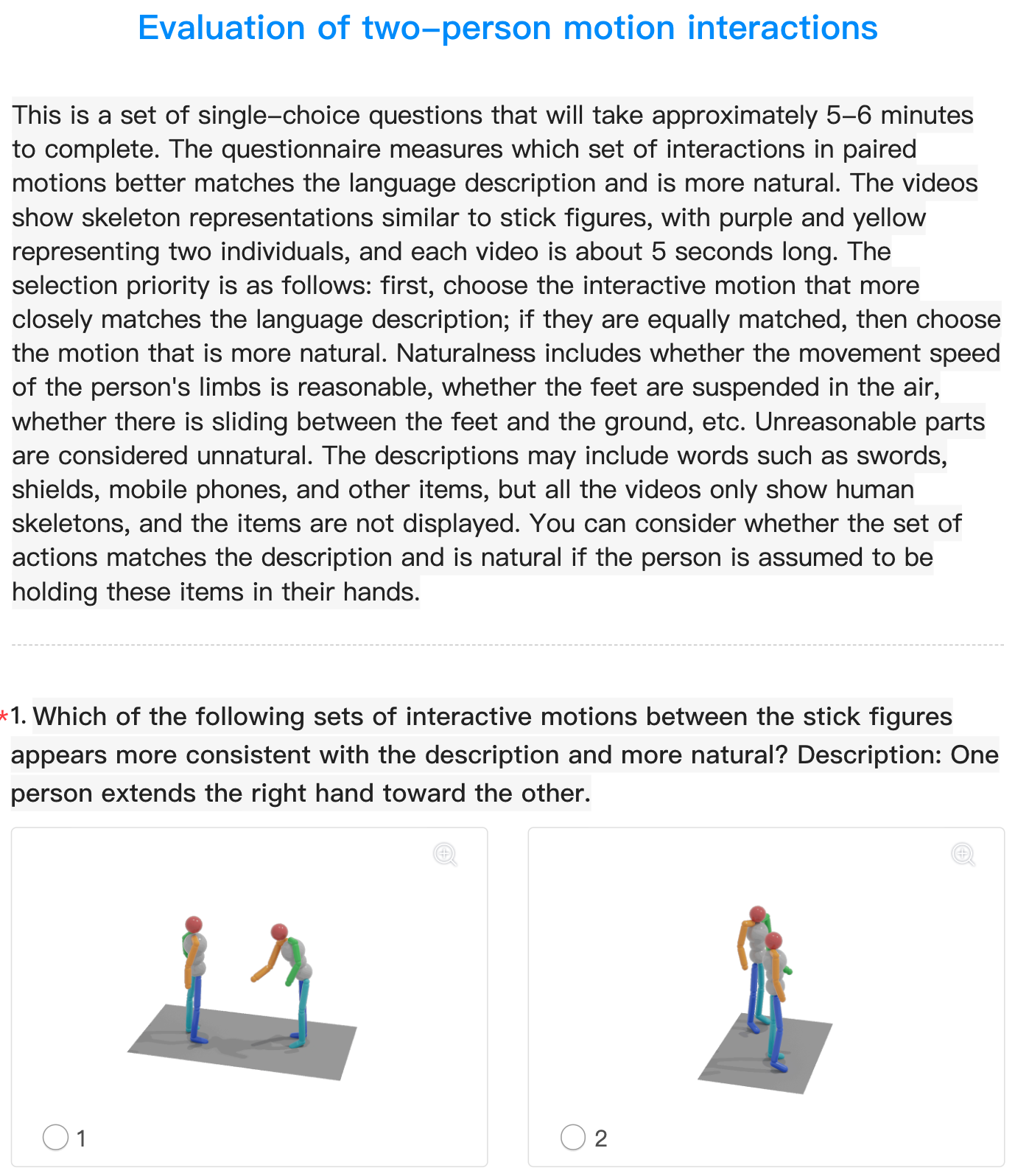}
    \vspace{-1em}
       \caption{\bf Example of the questionnaire of user-study.}
    \label{fig:user-study}
    
    \vspace{-1em}
\end{figure}

In the user study, our method generates 50 samples from the contact plans collected from LLM-planner. We also use the original interaction description to generate two-person interactions from ComMDM in PriorMDM~\cite{shafir2023human}. In Fig.~\ref{fig:user-study}, we show our designed questionnaire's evaluation instructions and the first question as an example. Each questionnaire has 19 single choice questions randomly sampled from all samples. In the folder named `user-study-videos', we provide 25 videos sampled from our Intercontrol and PriorMDM for reference.

\subsection{Details of Evaluation Metrics}
\label{sec:details_metrics}
Here we select some descriptions for metrics used to evaluate controllable motion generation methods from HumanML3D~\cite{guo2022t2m} and GMD~\cite{karunratanakul2023guided} to save reader's time.

{\bf Semantic-level Evaluation Metrics from HumanML3D~\cite{guo2022t2m}:} Frechet Inception Distance (FID), diversity and multi-modality. For quantitative evaluation, a motion feature extractor and text feature extractor is trained under contrastive loss to produce geometrically close feature vectors for matched text-motion pairs, and vice versa. Further explanations of aforementioned metrics as well as the specific textual and motion feature extractor are relegated to the supplementary file due to space limit. In addition, the R-precision and MultiModal distance are proposed in this work as complementary metrics, as follows. Consider R-precision: for each generated motion, its ground-truth text description and 31 randomly selected mismatched descriptions from the test set form a description pool. This is followed by calculating and ranking the Euclidean distances between the motion feature and the text feature of each description in the pool. We then count the average accuracy at top-1, top-2 and top-3 places. The ground truth entry falling into the top-k candidates is treated as successful retrieval, otherwise it fails. Meanwhile, MultiModal distance is computed as the average Euclidean distance between the motion feature of each generated motion and the text feature of its corresponding description in test set.

{\bf Spatial-level Evaluation Metrics from GMD~\cite{karunratanakul2023guided}:} We use Trajectory diversity, Trajectory error, Location error, and Average error of keyframe locations. Trajectory diversity measures the root mean square distance of each location of each motion step from the average location of that motion step across multiple samples with the same settings. Trajectory error is the ratio of unsuccessful trajectories, defined as those with any keyframe location error exceeding a threshold. Location error is the ratio of keyframe locations that are not reached within a threshold distance. Average error measures the mean distance between the generated motion locations and the keyframe locations measured at the keyframe motion steps.

\begin{table*}[t]
\small
  \begin{center}
  \caption{\textbf{Detailed prompting example of the LLM Planner.}}
  \label{tab:detailed_prompt_example} 
    \begin{tabular}{p{400pt}}
    \toprule
    Input \\
    \midrule
Instruction: two people greet each other with a handshake, while holding their cards in the left hand.\\
Given the instruction, generate 10 task plans according to the following background information, rules, and examples. Each task plan should completely reflect an entire process of actions described in the instruction.\\
 $[$start of background Information $[$\\
Human has JOINTS:  $[$'pelvis', 'left\_hip', 'right\_hip', 'left\_knee', 'right\_knee', 'left\_ankle', 'right\_ankle', 'left\_foot', 'right\_foot', 'neck', 'left\_collar', 'right\_collar', 'head', 'left\_shoulder', 'right\_shoulder', 'left\_elbow', 'right\_elbow', 'left\_wrist', 'right\_wrist' $[$.\\

The total number of TIME-STEPS of human motion is 99, the frame-per-second of motion is 20.\\

The provided text instruction is describing two people performing some actions containing human joint contacts.\\

The height of all people is 1.8 meters, the arm length is 0.6 meters, and the leg length is 0.9 meters.\\

Two people are 2 meters away at the beginning (i.e., TIME-STEPS=0).\\
$[$end of background Information$]$\\
$[$start of rules$]$\\
1. Each task plan should be composite into detailed steps.\\
2. Each step should contain meaningful joint-joint pairs.\\
3. Each joint-joint pair should be formatted into \{JOINT, JOINT, TIME-STEP, TIME-STEP, CONTACT TYPE, DISTANCE\}. JOINT should be replaced by JOINT in the background information. IMPORTANT: The first JOINT belongs to person 1, and the second JOINT belongs to person 2. Each joint-joint pair represents a contact of a joint of person 1 and a joint of person 2. The first TIME-STEP is the start frame number of contact, and the second TIME-STEP is the end frame number of contact. CONTACT TYPE should be selected from \{contact, avoid\}, DISTANCE should be a float number representing how many meters should be the distance of two joints in the joint-joint pair. For $[$CONTACT TYPE: contact$]$, the distance of two joints should be SMALLER than the DISTANCE; for $[$CONTACT TYPE: avoid$]$, the distance of two joints should be LARGER than the DISTANCE. IMPORTANT: Consider the transition of contact types, leave time-steps more than 20 frames without any joint-joint pair between different contact types. Use small DISTANCE variance between different contact types: for the joint-joint pairs that are with $[$CONTACT TYPE: contact$]$, do NOT use DISTANCE larger than 0.5m in the following $[$CONTACT TYPE: avoid$]$; for the joint-joint pairs that are with $[$CONTACT TYPE: contact$]$, use $[$CONTACT TYPE: avoid$]$ after 20 frames; for the joint-joint pairs that are with $[$CONTACT TYPE: avoid$]$, use NO joint pairs for 20 frames if the following CONTACT TYPE is contact. Try to not over-use $[$CONTACT TYPE: avoid$]$: if there is no explicit semantics of being far away, just do not use joint-joint pair in that frames; if there is explicit semantics of being far away, then use joint-joint pair with $[$CONTACT TYPE: avoid$]$.\\
4. Consider which JOINT will be interacted when two people perform the action described in the text instruction. Translate the text instruction to be steps of joint-joint pairs. Do not include extra joint-joint pairs that is unrelated to the text instruction. IMPORTANT: make joint-joint pairs in different task plans diverse in TIME-STEPS and JOINTs. Each joint-joint contact pairs should be lasting from 3 to 10 frames.\\
5. Be plausible. Do not generate uncommon interactions. Generate plausible interaction time-steps, and consider the velocity of human motions.\\
6. Use one sentence to describe what action should person 1 do and one sentence to describe what action should person 2 do according to the text instruction at the beginning of the task plan. IMPORTANT: the sentence starts from 'text 1:' describing the action of person 1 from the perspective of person 1 and the sentence starts from 'text 2:' describing the action of person 2 from the perspective of person 2. Sentences should NOT contain words like 'person 1' or 'person 2', use 'a person' to refer to himself in the sentence and 'others' to refer to others. \\
7. The steps in the task plan are for both two people. Use one set of steps to describe both two people. The first JOINT belongs to person 1, and the second JOINT belongs to person 2.\\
8. IMPORTANT: Do NOT add explanations for the steps in task plans. Each step only have one joint-joint pairs. \\

$[$end of rules$]$
\\
$[$start of an example$]$\\
Instruction: two people greet each other with a handshake, while holding their cards in the left hand.\\

$[$Start of Plan 1$]$\\
Text 1: a person make a handshake with others using his right wrist, while holding his cards in the left wrist.\\

Text 2: a person make a handshake with others using his right wrist, while holding his cards in the left wrist.\\

Step 1: \{right wrist, right wrist, 0, 10, avoid, 0.3\}\\

Step 2: \{right wrist, right wrist, 50, 60, contact, 0.05\}\\

Step 3: \{right wrist, right wrist, 90, 100, avoid, 0.3\}\\

$[$End of Plan 1$]$\\

$[$end of an example$]$\\
    \bottomrule
    \end{tabular}
  \end{center}
\end{table*}

\begin{table}[t]
\small
  \begin{center}
  \caption{\textbf{Example of the LLM generated task plans.}}
  \label{tab:task_plan} 
    \begin{tabular}{p{235pt}}
    \toprule
    LLM Generated Task Plans (only show 5 due to page limit) \\
    \midrule
    Instructions: The first fencer lunges at the second, who parries the attack and counters with a successful strike to the head. \\
    \midrule
$[$Start of Plan 1$]$\\
Text 1: A person lunges towards another with his right foot.\\
Text 2: A person parries the lunged attack while preparing to counter.\\
Step 1: \{right\_foot, left\_knee, 5, 10, contact, 0.3\}\\
Step 2: \{right\_wrist, left\_collar, 20, 30, avoid, 0.3\}\\
Step 3: \{left\_elbow, head, 70, 80, contact, 0.05\}\\
$[$End of Plan 1$]$\\
\\
$[$Start of Plan 2$]$\\
Text 1: A person lunges at the other person with his right foot.\\
Text 2: A person blocks the lunged attack.\\
Step 1: \{right\_foot, left\_ankle, 3, 10, contact, 0.2\}\\
Step 2: \{right\_wrist, right\_collar, 20, 30, avoid, 0.25\}\\
Step 3: \{left\_wrist, head, 70, 79, contact, 0.02\}\\
$[$End of Plan 2$]$\\
\\
$[$Start of Plan 3$]$\\
Text 1: A person takes a lunge step towards another.\\
Text 2: A person parries the attack and counters.\\
Step 1: \{right\_foot, right\_knee, 7, 14, contact, 0.3\}\\
Step 2: \{left\_wrist, right\_collar, 22, 30, avoid, 0.25\}\\
Step 3: \{right\_wrist, head, 69, 77, contact, 0.03\} \\
$[$End of Plan 3$]$\\
\\
$[$Start of Plan 4$]$\\
Text 1: A person lunges northerly towards another with his left foot.\\
Text 2: A person parries the attack and prepares a counterattack.\\
Step 1: \{left\_foot, right\_ankle, 6, 10, contact, 0.35\}\\
Step 2: \{left\_wrist, left\_collar, 22, 30, avoid, 0.28\}\\
Step 3: \{right\_elbow, head, 71, 80, contact, 0.05\}\\
$[$End of Plan 4$]$\\
\\
$[$Start of Plan 5$]$\\
Text 1: A person lunges at another using his right foot.\\
Text 2: A person deflects the approaching lunge and immediately counters.\\
Step 1: \{right\_foot, left\_knee, 4, 12, contact, 0.31\}\\
Step 2: \{left\_wrist, right\_shoulder, 20, 30, avoid, 0.3\}\\
Step 3: \{right\_wrist, head, 73, 81, contact, 0.05\}\\
$[$End of Plan 5$]$\\
    \bottomrule
    \end{tabular}
  \end{center}
\end{table}

\begin{table}[t]
\small
  \begin{center}
  \caption{\textbf{Example of processed json file from task plans generated by LLM.}}
  \label{tab:json} 
    \begin{tabular}{p{235pt}}
    \toprule
    Processed Json format, (only show 3 due to page limit) \\
    \midrule
    Format of 'steps': $[$ $[$index of contact joint of person 1, index of contact joint of person 2, \\ start frame, end frame, contact type (contact = 1, avoidance = 0), desired distance (unit as meter) $]$, ..., $]$. \\
    \midrule
$[$\\
\qquad\{\\
\qquad\qquad"text\_person1": "A person lunges towards another with his right foot.",\\
\qquad\qquad"text\_person2": "A person parries the lunged attack while preparing to counter.",\\
\qquad\qquad"steps": $[$\\
\qquad\qquad\qquad$[$11, 4, 5, 10, 1, 0.3$]$,\\
\qquad\qquad\qquad$[$21, 13, 30, 40, 0, 0.3$]$,\\
\qquad\qquad\qquad$[$18, 15, 70, 80, 1, 0.05$]$\\
\qquad\qquad$]$\\
\qquad\},\\
\qquad\{\\
\qquad\qquad"text\_person1": "A person lunges at the other person with his right foot.",\\
\qquad\qquad"text\_person2": "A person blocks the lunged attack.",\\
\qquad\qquad"steps": $[$\\
\qquad\qquad\qquad$[$11, 7, 3, 10, 1, 0.2$]$,\\
\qquad\qquad\qquad$[$21, 14, 30, 40, 0, 0.25$]$,\\
\qquad\qquad\qquad$[$20, 15, 70, 79, 1, 0.02$]$\\
\qquad\qquad$]$\\
\qquad\},\\
\qquad\{\\
\qquad\qquad"text\_person1": "A person takes a lunge step towards another.",\\
\qquad\qquad"text\_person2": "A person parries the attack and counters.",\\
\qquad\qquad"steps": $[$\\
\qquad\qquad\qquad$[$11, 5, 7, 14, 1, 0.3$]$,\\
\qquad\qquad\qquad$[$20, 14, 34, 42, 0, 0.25$]$,\\
\qquad\qquad\qquad$[$21, 15, 69, 77, 1, 0.03$]$\\
\qquad\qquad$]$\\
\qquad\}\\
$]$\\
    \bottomrule
    \end{tabular}
  \end{center}
\end{table}
\clearpage

%% file: neurips_2024.bbl
\begin{thebibliography}{73}
\providecommand{\natexlab}[1]{#1}
\providecommand{\url}[1]{\texttt{#1}}
\expandafter\ifx\csname urlstyle\endcsname\relax
  \providecommand{\doi}[1]{doi: #1}\else
  \providecommand{\doi}{doi: \begingroup \urlstyle{rm}\Url}\fi

\bibitem[CMU()]{CMU}
Cmu graphics lab motion capture database.

\bibitem[Ahuja and Morency(2019)]{ahuja2019language2pose}
Chaitanya Ahuja and Louis-Philippe Morency.
\newblock Language2pose: Natural language grounded pose forecasting.
\newblock In \emph{3DV}. IEEE, 2019.

\bibitem[Ao et~al.(2022)Ao, Gao, Lou, Chen, and
  Liu]{DBLP:journals/tog/AoGLCL22}
Tenglong Ao, Qingzhe Gao, Yuke Lou, Baoquan Chen, and Libin Liu.
\newblock Rhythmic gesticulator: Rhythm-aware co-speech gesture synthesis with
  hierarchical neural embeddings.
\newblock \emph{{ACM} Trans. Graph.}, 2022.

\bibitem[Bhattacharya et~al.(2021)Bhattacharya, Rewkowski, Banerjee, Guhan,
  Bera, and Manocha]{bhattacharya2021text2gestures}
Uttaran Bhattacharya, Nicholas Rewkowski, Abhishek Banerjee, Pooja Guhan,
  Aniket Bera, and Dinesh Manocha.
\newblock Text2gestures: A transformer-based network for generating emotive
  body gestures for virtual agents.
\newblock In \emph{VR}. IEEE, 2021.

\bibitem[Brown et~al.(2020)Brown, Mann, Ryder, Subbiah, Kaplan, Dhariwal,
  Neelakantan, Shyam, Sastry, Askell, Agarwal, Herbert{-}Voss, Krueger,
  Henighan, Child, Ramesh, Ziegler, Wu, Winter, Hesse, Chen, Sigler, Litwin,
  Gray, Chess, Clark, Berner, McCandlish, Radford, Sutskever, and
  Amodei]{DBLP:conf/nips/BrownMRSKDNSSAA20}
Tom~B. Brown, Benjamin Mann, Nick Ryder, Melanie Subbiah, Jared Kaplan,
  Prafulla Dhariwal, Arvind Neelakantan, Pranav Shyam, Girish Sastry, Amanda
  Askell, Sandhini Agarwal, Ariel Herbert{-}Voss, Gretchen Krueger, Tom
  Henighan, Rewon Child, Aditya Ramesh, Daniel~M. Ziegler, Jeffrey Wu, Clemens
  Winter, Christopher Hesse, Mark Chen, Eric Sigler, Mateusz Litwin, Scott
  Gray, Benjamin Chess, Jack Clark, Christopher Berner, Sam McCandlish, Alec
  Radford, Ilya Sutskever, and Dario Amodei.
\newblock Language models are few-shot learners.
\newblock In \emph{NeurIPS}, 2020.

\bibitem[Chen et~al.(2023)Chen, Jiang, Liu, Huang, Fu, Chen, and
  Yu]{DBLP:conf/cvpr/ChenJLHFCY23}
Xin Chen, Biao Jiang, Wen Liu, Zilong Huang, Bin Fu, Tao Chen, and Gang Yu.
\newblock Executing your commands via motion diffusion in latent space.
\newblock In \emph{{CVPR}}, 2023.

\bibitem[Choi et~al.(2021)Choi, Kim, Jeong, Gwon, and
  Yoon]{DBLP:conf/iccv/ChoiKJGY21}
Jooyoung Choi, Sungwon Kim, Yonghyun Jeong, Youngjune Gwon, and Sungroh Yoon.
\newblock {ILVR:} conditioning method for denoising diffusion probabilistic
  models.
\newblock In \emph{{ICCV}}, 2021.

\bibitem[Chung et~al.(2022)Chung, Sim, Ryu, and Ye]{DBLP:conf/nips/ChungSRY22}
Hyungjin Chung, Byeongsu Sim, Dohoon Ryu, and Jong~Chul Ye.
\newblock Improving diffusion models for inverse problems using manifold
  constraints.
\newblock In \emph{NeurIPS}, 2022.

\bibitem[Dhariwal and Nichol(2021)]{DBLP:conf/nips/DhariwalN21}
Prafulla Dhariwal and Alexander~Quinn Nichol.
\newblock Diffusion models beat gans on image synthesis.
\newblock In \emph{NeurIPS}, 2021.

\bibitem[Duan et~al.(2021)Duan, Shi, Zou, Lin, Qian, Zhang, and
  Yuan]{duan2021single}
Yinglin Duan, Tianyang Shi, Zhengxia Zou, Yenan Lin, Zhehui Qian, Bohan Zhang,
  and Yi Yuan.
\newblock Single-shot motion completion with transformer.
\newblock \emph{arXiv preprint arXiv:2103.00776}, 2021.

\bibitem[Esser et~al.(2023)Esser, Chiu, Atighehchian, Granskog, and
  Germanidis]{esser2023structure}
Patrick Esser, Johnathan Chiu, Parmida Atighehchian, Jonathan Granskog, and
  Anastasis Germanidis.
\newblock Structure and content-guided video synthesis with diffusion models.
\newblock In \emph{ICCV}, 2023.

\bibitem[Ghosh et~al.(2023)Ghosh, Dabral, Golyanik, Theobalt, and
  Slusallek]{DBLP:journals/cgf/GhoshDGTS23}
Anindita Ghosh, Rishabh Dabral, Vladislav Golyanik, Christian Theobalt, and
  Philipp Slusallek.
\newblock Imos: Intent-driven full-body motion synthesis for human-object
  interactions.
\newblock \emph{Comput. Graph. Forum}, 2023.

\bibitem[Guo et~al.(2020)Guo, Zuo, Wang, Zou, Sun, Deng, Gong, and
  Cheng]{guo2020action2motion}
Chuan Guo, Xinxin Zuo, Sen Wang, Shihao Zou, Qingyao Sun, Annan Deng, Minglun
  Gong, and Li Cheng.
\newblock Action2motion: Conditioned generation of 3d human motions.
\newblock In \emph{ACM MM}, 2020.

\bibitem[Guo et~al.(2022{\natexlab{a}})Guo, Zou, Zuo, Wang, Ji, Li, and
  Cheng]{guo2022t2m}
Chuan Guo, Shihao Zou, Xinxin Zuo, Sen Wang, Wei Ji, Xingyu Li, and Li Cheng.
\newblock Generating diverse and natural 3d human motions from text.
\newblock In \emph{CVPR}, 2022{\natexlab{a}}.

\bibitem[Guo et~al.(2022{\natexlab{b}})Guo, Zuo, Wang, and Cheng]{guo2022tm2t}
Chuan Guo, Xinxin Zuo, Sen Wang, and Li Cheng.
\newblock Tm2t: Stochastic and tokenized modeling for the reciprocal generation
  of 3d human motions and texts.
\newblock In \emph{ECCV}, 2022{\natexlab{b}}.

\bibitem[Guo et~al.(2022{\natexlab{c}})Guo, Bie, Alameda{-}Pineda, and
  Moreno{-}Noguer]{DBLP:conf/cvpr/GuoBAM22}
Wen Guo, Xiaoyu Bie, Xavier Alameda{-}Pineda, and Francesc Moreno{-}Noguer.
\newblock Multi-person extreme motion prediction.
\newblock In \emph{{CVPR}}, 2022{\natexlab{c}}.

\bibitem[Guo et~al.(2023)Guo, Yang, Rao, Wang, Qiao, Lin, and
  Dai]{guo2023animatediff}
Yuwei Guo, Ceyuan Yang, Anyi Rao, Yaohui Wang, Yu Qiao, Dahua Lin, and Bo Dai.
\newblock Animatediff: Animate your personalized text-to-image diffusion models
  without specific tuning.
\newblock \emph{arXiv preprint arXiv:2307.04725}, 2023.

\bibitem[Habibie et~al.(2022)Habibie, Elgharib, Sarkar, Abdullah, Nyatsanga,
  Neff, and Theobalt]{DBLP:conf/siggraph/HabibieESANNT22}
Ikhsanul Habibie, Mohamed Elgharib, Kripasindhu Sarkar, Ahsan Abdullah,
  Simbarashe Nyatsanga, Michael Neff, and Christian Theobalt.
\newblock A motion matching-based framework for controllable gesture synthesis
  from speech.
\newblock In \emph{{SIGGRAPH} (Conference Paper Track)}, 2022.

\bibitem[Harvey et~al.(2020)Harvey, Yurick, Nowrouzezahrai, and
  Pal]{harvey2020robust}
F{\'e}lix~G Harvey, Mike Yurick, Derek Nowrouzezahrai, and Christopher Pal.
\newblock Robust motion in-betweening.
\newblock \emph{ACM Transactions on Graphics (TOG)}, 2020.

\bibitem[Hassan et~al.(2021)Hassan, Ceylan, Villegas, Saito, Yang, Zhou, and
  Black]{DBLP:conf/iccv/HassanCVSYZB21}
Mohamed Hassan, Duygu Ceylan, Ruben Villegas, Jun Saito, Jimei Yang, Yi Zhou,
  and Michael~J. Black.
\newblock Stochastic scene-aware motion prediction.
\newblock In \emph{{ICCV}}, 2021.

\bibitem[He et~al.(2016)He, Zhang, Ren, and Sun]{DBLP:conf/cvpr/HeZRS16}
Kaiming He, Xiangyu Zhang, Shaoqing Ren, and Jian Sun.
\newblock Deep residual learning for image recognition.
\newblock In \emph{{CVPR}}, 2016.

\bibitem[Ho and Salimans(2022)]{ho2022classifier}
Jonathan Ho and Tim Salimans.
\newblock Classifier-free diffusion guidance.
\newblock \emph{arXiv preprint arXiv:2207.12598}, 2022.

\bibitem[Ho et~al.(2020)Ho, Jain, and Abbeel]{DBLP:conf/nips/HoJA20}
Jonathan Ho, Ajay Jain, and Pieter Abbeel.
\newblock Denoising diffusion probabilistic models.
\newblock In \emph{NeurIPS}, 2020.

\bibitem[Ho et~al.(2022)Ho, Chan, Saharia, Whang, Gao, Gritsenko, Kingma,
  Poole, Norouzi, Fleet, et~al.]{ho2022imagen}
Jonathan Ho, William Chan, Chitwan Saharia, Jay Whang, Ruiqi Gao, Alexey
  Gritsenko, Diederik~P Kingma, Ben Poole, Mohammad Norouzi, David~J Fleet,
  et~al.
\newblock Imagen video: High definition video generation with diffusion models.
\newblock \emph{arXiv preprint arXiv:2210.02303}, 2022.

\bibitem[Jiang et~al.(2023)Jiang, Chen, Liu, Yu, Yu, and
  Chen]{jiang2023motiongpt}
Biao Jiang, Xin Chen, Wen Liu, Jingyi Yu, Gang Yu, and Tao Chen.
\newblock Motiongpt: Human motion as a foreign language.
\newblock \emph{arXiv preprint arXiv:2306.14795}, 2023.

\bibitem[Jiang et~al.(2022)Jiang, Liu, Cao, Cui, Chen, Wang, Zhu, and
  Huang]{jiang2022chairs}
Nan Jiang, Tengyu Liu, Zhexuan Cao, Jieming Cui, Yixin Chen, He Wang, Yixin
  Zhu, and Siyuan Huang.
\newblock Chairs: Towards full-body articulated human-object interaction.
\newblock \emph{arXiv preprint arXiv:2212.10621}, 2022.

\bibitem[Karunratanakul et~al.(2023)Karunratanakul, Preechakul, Suwajanakorn,
  and Tang]{karunratanakul2023guided}
Korrawe Karunratanakul, Konpat Preechakul, Supasorn Suwajanakorn, and Siyu
  Tang.
\newblock Guided motion diffusion for controllable human motion synthesis.
\newblock In \emph{CVPR}, 2023.

\bibitem[Kaufmann et~al.(2020)Kaufmann, Aksan, Song, Pece, Ziegler, and
  Hilliges]{DBLP:conf/3dim/KaufmannA0PZH20}
Manuel Kaufmann, Emre Aksan, Jie Song, Fabrizio Pece, Remo Ziegler, and Otmar
  Hilliges.
\newblock Convolutional autoencoders for human motion infilling.
\newblock In \emph{3DV}, 2020.

\bibitem[Kim et~al.(2021)Kim, Seol, and Kwon]{DBLP:journals/jvca/KimSK21}
Jongmin Kim, Yeongho Seol, and Taesoo Kwon.
\newblock Interactive multi-character motion retargeting.
\newblock \emph{Comput. Animat. Virtual Worlds}, 2021.

\bibitem[Kim et~al.(2023)Kim, Kim, and Choi]{DBLP:conf/aaai/KimKC23}
Jihoon Kim, Jiseob Kim, and Sungjoon Choi.
\newblock {FLAME:} free-form language-based motion synthesis {\&} editing.
\newblock In \emph{{AAAI}}, 2023.

\bibitem[Kingma and Welling(2014)]{DBLP:journals/corr/KingmaW13}
Diederik~P. Kingma and Max Welling.
\newblock Auto-encoding variational bayes.
\newblock In \emph{{ICLR}}, 2014.

\bibitem[Kong et~al.(2020)Kong, Ping, Huang, Zhao, and
  Catanzaro]{kong2020diffwave}
Zhifeng Kong, Wei Ping, Jiaji Huang, Kexin Zhao, and Bryan Catanzaro.
\newblock Diffwave: A versatile diffusion model for audio synthesis.
\newblock \emph{arXiv preprint arXiv:2009.09761}, 2020.

\bibitem[Kulkarni et~al.(2023)Kulkarni, Rempe, Genova, Kundu, Johnson, Fouhey,
  and Guibas]{kulkarni2023nifty}
Nilesh Kulkarni, Davis Rempe, Kyle Genova, Abhijit Kundu, Justin Johnson, David
  Fouhey, and Leonidas Guibas.
\newblock Nifty: Neural object interaction fields for guided human motion
  synthesis.
\newblock \emph{arXiv preprint arXiv:2307.07511}, 2023.

\bibitem[Li et~al.(2022)Li, Zhao, Zhelun, and Sheng]{li2022danceformer}
Buyu Li, Yongchi Zhao, Shi Zhelun, and Lu Sheng.
\newblock Danceformer: Music conditioned 3d dance generation with parametric
  motion transformer.
\newblock In \emph{AAAI}, 2022.

\bibitem[Li et~al.(2021)Li, Yang, Ross, and Kanazawa]{li2021ai}
Ruilong Li, Shan Yang, David~A Ross, and Angjoo Kanazawa.
\newblock Ai choreographer: Music conditioned 3d dance generation with aist++.
\newblock In \emph{ICCV}, 2021.

\bibitem[Liang et~al.(2023)Liang, Zhang, Li, Yu, and Xu]{liang2023intergen}
Han Liang, Wenqian Zhang, Wenxuan Li, Jingyi Yu, and Lan Xu.
\newblock Intergen: Diffusion-based multi-human motion generation under complex
  interactions.
\newblock \emph{arXiv preprint arXiv:2304.05684}, 2023.

\bibitem[Liu and Nocedal(1989)]{DBLP:journals/mp/LiuN89}
Dong~C. Liu and Jorge Nocedal.
\newblock On the limited memory {BFGS} method for large scale optimization.
\newblock \emph{Math. Program.}, 1989.

\bibitem[Loper et~al.(2015)Loper, Mahmood, Romero, Pons{-}Moll, and
  Black]{DBLP:journals/tog/LoperM0PB15}
Matthew Loper, Naureen Mahmood, Javier Romero, Gerard Pons{-}Moll, and
  Michael~J. Black.
\newblock {SMPL:} a skinned multi-person linear model.
\newblock \emph{{ACM} Trans. Graph.}, 2015.

\bibitem[Loshchilov and Hutter(2019)]{DBLP:conf/iclr/LoshchilovH19}
Ilya Loshchilov and Frank Hutter.
\newblock Decoupled weight decay regularization.
\newblock In \emph{{ICLR} (Poster)}, 2019.

\bibitem[Luo et~al.(2023)Luo, Cao, Winkler, Kitani, and
  Xu]{DBLP:conf/iccv/0002CWKX23}
Zhengyi Luo, Jinkun Cao, Alexander Winkler, Kris Kitani, and Weipeng Xu.
\newblock Perpetual humanoid control for real-time simulated avatars.
\newblock In \emph{{ICCV}}, 2023.

\bibitem[Mahmood et~al.(2019)Mahmood, Ghorbani, Troje, Pons-Moll, and
  Black]{mahmood2019amass}
Naureen Mahmood, Nima Ghorbani, Nikolaus~F Troje, Gerard Pons-Moll, and
  Michael~J Black.
\newblock Amass: Archive of motion capture as surface shapes.
\newblock In \emph{ICCV}, 2019.

\bibitem[Mehta et~al.(2018)Mehta, Sotnychenko, Mueller, Xu, Sridhar,
  Pons{-}Moll, and Theobalt]{DBLP:conf/3dim/MehtaSMX0PT18}
Dushyant Mehta, Oleksandr Sotnychenko, Franziska Mueller, Weipeng Xu, Srinath
  Sridhar, Gerard Pons{-}Moll, and Christian Theobalt.
\newblock Single-shot multi-person 3d pose estimation from monocular {RGB}.
\newblock In \emph{3DV}, 2018.

\bibitem[OpenAI(2023)]{DBLP:journals/corr/abs-2303-08774}
OpenAI.
\newblock {GPT-4} technical report.
\newblock \emph{arXiv preprint arXiv:2303.08774}, 2023.

\bibitem[Pavlakos et~al.(2019)Pavlakos, Choutas, Ghorbani, Bolkart, Osman,
  Tzionas, and Black]{DBLP:conf/cvpr/PavlakosCGBOTB19}
Georgios Pavlakos, Vasileios Choutas, Nima Ghorbani, Timo Bolkart, Ahmed A.~A.
  Osman, Dimitrios Tzionas, and Michael~J. Black.
\newblock Expressive body capture: 3d hands, face, and body from a single
  image.
\newblock In \emph{{CVPR}}, 2019.

\bibitem[Petrovich et~al.(2021)Petrovich, Black, and
  Varol]{petrovich2021action}
Mathis Petrovich, Michael~J Black, and G{\"u}l Varol.
\newblock Action-conditioned 3d human motion synthesis with transformer vae.
\newblock In \emph{ICCV}, 2021.

\bibitem[Petrovich et~al.(2022)Petrovich, Black, and Varol]{petrovich2022temos}
Mathis Petrovich, Michael~J Black, and G{\"u}l Varol.
\newblock Temos: Generating diverse human motions from textual descriptions.
\newblock In \emph{ECCV}, 2022.

\bibitem[Plappert et~al.(2016)Plappert, Mandery, and
  Asfour]{DBLP:journals/bigdata/PlappertMA16}
Matthias Plappert, Christian Mandery, and Tamim Asfour.
\newblock The {KIT} motion-language dataset.
\newblock \emph{Big Data}, 2016.

\bibitem[Radford et~al.(2021)Radford, Kim, Hallacy, Ramesh, Goh, Agarwal,
  Sastry, Askell, Mishkin, Clark, Krueger, and
  Sutskever]{DBLP:conf/icml/RadfordKHRGASAM21}
Alec Radford, Jong~Wook Kim, Chris Hallacy, Aditya Ramesh, Gabriel Goh,
  Sandhini Agarwal, Girish Sastry, Amanda Askell, Pamela Mishkin, Jack Clark,
  Gretchen Krueger, and Ilya Sutskever.
\newblock Learning transferable visual models from natural language
  supervision.
\newblock In \emph{{ICML}}, 2021.

\bibitem[Rempe et~al.(2023)Rempe, Luo, Peng, Yuan, Kitani, Kreis, Fidler, and
  Litany]{DBLP:conf/cvpr/Rempe0P0KKFL23}
Davis Rempe, Zhengyi Luo, Xue~Bin Peng, Ye Yuan, Kris Kitani, Karsten Kreis,
  Sanja Fidler, and Or Litany.
\newblock Trace and pace: Controllable pedestrian animation via guided
  trajectory diffusion.
\newblock In \emph{{CVPR}}, 2023.

\bibitem[Rombach et~al.(2022)Rombach, Blattmann, Lorenz, Esser, and
  Ommer]{DBLP:conf/cvpr/RombachBLEO22}
Robin Rombach, Andreas Blattmann, Dominik Lorenz, Patrick Esser, and
  Bj{\"{o}}rn Ommer.
\newblock High-resolution image synthesis with latent diffusion models.
\newblock In \emph{{CVPR}}, 2022.

\bibitem[Shafir et~al.(2023)Shafir, Tevet, Kapon, and Bermano]{shafir2023human}
Yonatan Shafir, Guy Tevet, Roy Kapon, and Amit~H Bermano.
\newblock Human motion diffusion as a generative prior.
\newblock \emph{arXiv preprint arXiv:2303.01418}, 2023.

\bibitem[Song et~al.(2023)Song, Zhang, Yin, Mardani, Liu, Kautz, Chen, and
  Vahdat]{DBLP:conf/icml/SongZYM0KCV23}
Jiaming Song, Qinsheng Zhang, Hongxu Yin, Morteza Mardani, Ming{-}Yu Liu, Jan
  Kautz, Yongxin Chen, and Arash Vahdat.
\newblock Loss-guided diffusion models for plug-and-play controllable
  generation.
\newblock In \emph{{ICML}}, 2023.

\bibitem[Song et~al.(2021)Song, Sohl{-}Dickstein, Kingma, Kumar, Ermon, and
  Poole]{DBLP:conf/iclr/0011SKKEP21}
Yang Song, Jascha Sohl{-}Dickstein, Diederik~P. Kingma, Abhishek Kumar, Stefano
  Ermon, and Ben Poole.
\newblock Score-based generative modeling through stochastic differential
  equations.
\newblock In \emph{{ICLR}}, 2021.

\bibitem[Starke et~al.(2019)Starke, Zhang, Komura, and
  Saito]{DBLP:journals/tog/StarkeZKS19}
Sebastian Starke, He Zhang, Taku Komura, and Jun Saito.
\newblock Neural state machine for character-scene interactions.
\newblock \emph{{ACM} Trans. Graph.}, 2019.

\bibitem[Tevet et~al.(2023)Tevet, Raab, Gordon, Shafir, Cohen{-}Or, and
  Bermano]{DBLP:conf/iclr/TevetRGSCB23}
Guy Tevet, Sigal Raab, Brian Gordon, Yonatan Shafir, Daniel Cohen{-}Or, and
  Amit~Haim Bermano.
\newblock Human motion diffusion model.
\newblock In \emph{{ICLR}}, 2023.

\bibitem[Tseng et~al.(2023)Tseng, Castellon, and Liu]{DBLP:conf/cvpr/TsengCL23}
Jonathan Tseng, Rodrigo Castellon, and C.~Karen Liu.
\newblock {EDGE:} editable dance generation from music.
\newblock In \emph{{CVPR}}, 2023.

\bibitem[Vaillant et~al.(2017)Vaillant, Bouyarmane, and
  Kheddar]{DBLP:journals/tvcg/VaillantBK17}
Joris Vaillant, Karim Bouyarmane, and Abderrahmane Kheddar.
\newblock Multi-character physical and behavioral interactions controller.
\newblock \emph{{IEEE} Trans. Vis. Comput. Graph.}, 2017.

\bibitem[Vaswani et~al.(2017)Vaswani, Shazeer, Parmar, Uszkoreit, Jones, Gomez,
  Kaiser, and Polosukhin]{DBLP:conf/nips/VaswaniSPUJGKP17}
Ashish Vaswani, Noam Shazeer, Niki Parmar, Jakob Uszkoreit, Llion Jones,
  Aidan~N. Gomez, Lukasz Kaiser, and Illia Polosukhin.
\newblock Attention is all you need.
\newblock In \emph{{NIPS}}, 2017.

\bibitem[von Marcard et~al.(2018)von Marcard, Henschel, Black, Rosenhahn, and
  Pons{-}Moll]{DBLP:conf/eccv/MarcardHBRP18}
Timo von Marcard, Roberto Henschel, Michael~J. Black, Bodo Rosenhahn, and
  Gerard Pons{-}Moll.
\newblock Recovering accurate 3d human pose in the wild using imus and a moving
  camera.
\newblock In \emph{{ECCV}}, 2018.

\bibitem[Wang et~al.(2021{\natexlab{a}})Wang, Xu, Narasimhan, and
  Wang]{DBLP:conf/nips/WangXNW21}
Jiashun Wang, Huazhe Xu, Medhini Narasimhan, and Xiaolong Wang.
\newblock Multi-person 3d motion prediction with multi-range transformers.
\newblock In \emph{NeurIPS}, 2021{\natexlab{a}}.

\bibitem[Wang et~al.(2021{\natexlab{b}})Wang, Xu, Xu, Liu, and
  Wang]{wang2021synthesizing}
Jiashun Wang, Huazhe Xu, Jingwei Xu, Sifei Liu, and Xiaolong Wang.
\newblock Synthesizing long-term 3d human motion and interaction in 3d scenes.
\newblock In \emph{CVPR}, 2021{\natexlab{b}}.

\bibitem[Wang et~al.(2021{\natexlab{c}})Wang, Yan, Dai, and
  Lin]{DBLP:conf/cvpr/WangYDL21}
Jingbo Wang, Sijie Yan, Bo Dai, and Dahua Lin.
\newblock Scene-aware generative network for human motion synthesis.
\newblock In \emph{{CVPR}}, 2021{\natexlab{c}}.

\bibitem[Wang et~al.(2022)Wang, Chen, Liu, Zhu, Liang, and
  Huang]{DBLP:conf/nips/WangCLZLH22}
Zan Wang, Yixin Chen, Tengyu Liu, Yixin Zhu, Wei Liang, and Siyuan Huang.
\newblock {HUMANISE:} language-conditioned human motion generation in 3d
  scenes.
\newblock In \emph{NeurIPS}, 2022.

\bibitem[Xiao et~al.(2023)Xiao, Wang, Wang, Cao, Zhang, Dai, Lin, and
  Pang]{xiao2023unified}
Zeqi Xiao, Tai Wang, Jingbo Wang, Jinkun Cao, Wenwei Zhang, Bo Dai, Dahua Lin,
  and Jiangmiao Pang.
\newblock Unified human-scene interaction via prompted chain-of-contacts.
\newblock \emph{arXiv preprint arXiv:2309.07918}, 2023.

\bibitem[Xie et~al.(2023)Xie, Jampani, Zhong, Sun, and
  Jiang]{xie2023omnicontrol}
Yiming Xie, Varun Jampani, Lei Zhong, Deqing Sun, and Huaizu Jiang.
\newblock Omnicontrol: Control any joint at any time for human motion
  generation.
\newblock \emph{arXiv preprint arXiv:2310.08580}, 2023.

\bibitem[Xu et~al.(2023{\natexlab{a}})Xu, Li, Wang, and Gui]{xu2023interdiff}
Sirui Xu, Zhengyuan Li, Yu-Xiong Wang, and Liang-Yan Gui.
\newblock Interdiff: Generating 3d human-object interactions with
  physics-informed diffusion.
\newblock In \emph{ICCV}, 2023{\natexlab{a}}.

\bibitem[Xu et~al.(2023{\natexlab{b}})Xu, Wang, and
  Gui]{DBLP:conf/iclr/0002WG23}
Sirui Xu, Yu{-}Xiong Wang, and Liangyan Gui.
\newblock Stochastic multi-person 3d motion forecasting.
\newblock In \emph{{ICLR}}, 2023{\natexlab{b}}.

\bibitem[Yuan et~al.(2023)Yuan, Song, Iqbal, Vahdat, and
  Kautz]{yuan2023physdiff}
Ye Yuan, Jiaming Song, Umar Iqbal, Arash Vahdat, and Jan Kautz.
\newblock Physdiff: Physics-guided human motion diffusion model.
\newblock In \emph{ICCV}, 2023.

\bibitem[Zhang et~al.(2023{\natexlab{a}})Zhang, Zhang, Cun, Zhang, Zhao, Lu,
  Shen, and Ying]{DBLP:conf/cvpr/ZhangZCZZLSY23}
Jianrong Zhang, Yangsong Zhang, Xiaodong Cun, Yong Zhang, Hongwei Zhao, Hongtao
  Lu, Xi Shen, and Shan Ying.
\newblock Generating human motion from textual descriptions with discrete
  representations.
\newblock In \emph{{CVPR}}, 2023{\natexlab{a}}.

\bibitem[Zhang et~al.(2023{\natexlab{b}})Zhang, Rao, and
  Agrawala]{zhang2023adding}
Lvmin Zhang, Anyi Rao, and Maneesh Agrawala.
\newblock Adding conditional control to text-to-image diffusion models.
\newblock In \emph{ICCV}, 2023{\natexlab{b}}.

\bibitem[Zhang et~al.(2022)Zhang, Cai, Pan, Hong, Guo, Yang, and
  Liu]{zhang2022motiondiffuse}
Mingyuan Zhang, Zhongang Cai, Liang Pan, Fangzhou Hong, Xinying Guo, Lei Yang,
  and Ziwei Liu.
\newblock Motiondiffuse: Text-driven human motion generation with diffusion
  model.
\newblock \emph{arXiv preprint arXiv:2208.15001}, 2022.

\bibitem[Zhang et~al.(2023{\natexlab{c}})Zhang, Gopinath, Ye, Hodgins, Turk,
  and Won]{DBLP:conf/siggraph/ZhangGYHTW23}
Yunbo Zhang, Deepak Gopinath, Yuting Ye, Jessica~K. Hodgins, Greg Turk, and
  Jungdam Won.
\newblock Simulation and retargeting of complex multi-character interactions.
\newblock In \emph{{SIGGRAPH} (Conference Paper Track)}, 2023{\natexlab{c}}.

\bibitem[Zhao et~al.(2023)Zhao, Zhang, Wang, Beeler, and
  Tang]{zhao2023synthesizing}
Kaifeng Zhao, Yan Zhang, Shaofei Wang, Thabo Beeler, and Siyu Tang.
\newblock Synthesizing diverse human motions in 3d indoor scenes.
\newblock \emph{arXiv preprint arXiv:2305.12411}, 2023.

\end{thebibliography}
